%% file: KnowledgeEdit.tex
\documentclass[acmsmall,screen]{acmart}

\usepackage{subcaption}
\usepackage{multirow}
\usepackage{tcolorbox}
\usepackage{xspace}
\usepackage{pifont}

\newcommand{\ie}{\emph{i.e.,}\xspace}

\newcommand{\cmark}{\ding{51}}%
\newcommand{\xmark}{\ding{55}}%

\AtBeginDocument{%
  }

\setcopyright{acmlicensed}
\copyrightyear{2025}
\acmYear{2025}
\acmDOI{XXXXXXX.XXXXXXX}

\acmJournal{JACM}
\acmVolume{1}
\acmNumber{1}
\acmArticle{1}
\acmMonth{3}

\begin{document}

\title{Knowledge Updating? No More Model Editing! Just Selective Contextual Reasoning}

\author{Guoxiu He}
\email{gxhe@fem.ecnu.edu.cn}
\orcid{0000-0002-1419-7495}

\author{Xin Song}
\email{xsong2023@stu.ecnu.edu.cn}
\orcid{0009-0009-3323-5846}
\affiliation{%
 \institution{School of Economics and Management, East China Normal University}
 \city{Shanghai}
 \country{China}
}

\author{Aixin Sun}
%\authornotemark[2]
\authornote{Corresponding author.}
\email{axsun@ntu.edu.sg}
\orcid{0000-0003-0764-4258}
\affiliation{%
 \institution{College of Computing and Data Science, Nanyang Technological University}
 \city{Singapore}
 \country{Singapore}
}

\begin{abstract}
As real-world knowledge evolves, the information embedded within large language models (LLMs) can become outdated, inadequate, or erroneous. Model editing has emerged as a prominent approach for updating LLMs' knowledge with minimal computational costs and parameter changes. This approach typically identifies and adjusts specific model parameters associated with newly acquired knowledge. However, existing methods often underestimate the adverse effects that parameter modifications can have on broadly distributed knowledge. More critically, post-edit LLMs frequently struggle with multi-hop reasoning and continuous knowledge updates. Although various studies have discussed these shortcomings, there is a lack of comprehensive evaluation. In this paper, we provide an evaluation of ten model editing methods along four dimensions: reliability, generalization, locality, and portability. Results confirm that all ten popular model editing methods show significant shortcomings across multiple dimensions, suggesting \textit{model editing is less promising}.  We then propose a straightforward method called \textit{Selective Contextual Reasoning} (\textbf{SCR}), for knowledge updating. SCR does not modify model parameters but harnesses LLM's inherent contextual reasoning capabilities utilizing the updated knowledge pieces. Under SCR, an LLM first assesses whether an incoming query falls within the scope of an external knowledge base. If it does, the relevant external knowledge texts are contextualized to enhance reasoning; otherwise, the query is answered directly.  We evaluate SCR  against the ten model editing methods on two counterfactual datasets with three backbone LLMs. Empirical results confirm the effectiveness and efficiency of contextual reasoning for knowledge updating. 
\end{abstract}




 % knowledge editing, sequential editing


\maketitle

\input{fullContent}

%\begin{acks}
%To Robert, for the bagels and explaining CMYK and color spaces.
%\end{acks}

\bibliographystyle{ACM-Reference-Format}
\bibliography{KnowledgeEdit}
\end{document}

%% file: fullContent.tex
%======================
\section{Introduction}
\label{sec:intro}
%======================

Large language models (LLMs) represent a significant breakthrough in artificial intelligence (AI) \cite{zeng2022glm,touvron2023llama,openai2023gpt}. By undergoing extensive pre-training on vast datasets \cite{brown2020language, ouyang2022training}, LLMs acquire a profound understanding of the world knowledge \cite{jiang2020can, alkhamissi2022review, zhang2023large} and demonstrate remarkable contextual reasoning abilities \cite{openai2023gpt, liu2023pre, lee2024reasoning}. Currently, LLMs predominantly lead the field of natural language processing (NLP) tasks \cite{kamalloo2023evaluating, yang2024harnessing}. However, the world is constantly changing. For instance, 
the president of a county may change after an election.
Consequently, pre-trained LLMs are prone to becoming outdated and erroneous as new information and trends emerge over time \cite{mousavi2024your,ji2023survey}.

\begin{figure}
    \centering
    \includegraphics[width=\textwidth]{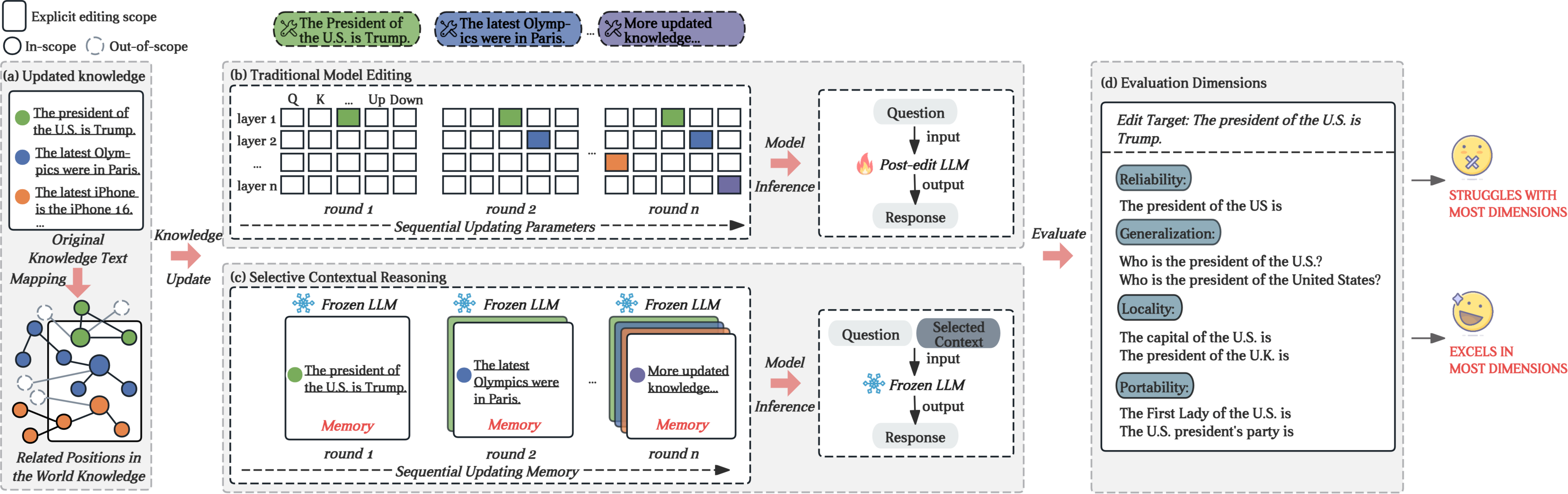}
    \caption{The pipeline of the knowledge updating. Given the collected knowledge dataset, the key differences between traditional model editing and our method lie in how they handle knowledge updates and perform inference on factual questions.
    During knowledge updating, traditional model editing relies on modifying internal parameters, whereas our approach updates an external memory. During model inferencing, traditional model editing tends to accumulate significant bias over multiple operations, often leading to instability or failure. In contrast, our method, which leverages selective contextual reasoning, explicitly incorporating updated knowledge into the context, which enables better performance on multi-hop questions.
    }	
    \label{fig:model_intro}
\end{figure}

To bridge this gap, the necessary knowledge pieces are typically collected and organized into textual data, as shown in the upper half of Fig. \ref{fig:model_intro} (a). However, re-training LLMs from scratch to incorporate the limited knowledge updates is both computationally expensive and time-consuming. Instead, full fine-tuning (FFT) \cite{lv2024full} enables continuing to adjust all parameters in a pre-trained LLM to adapt it to new knowledge more efficiently. Parameter-efficient fine-tuning (PEFT) \cite{ding2023parameter} takes this a step further by minimizing the number of parameters that need adjustment. Furthermore, \textit{model editing}, also known as \textit{knowledge editing} \cite{sinitsin2020editable, zhumodifying}, as shown in Fig. \ref{fig:model_intro} (b), has garnered significant attention to align LLMs with the up-to-date knowledge through minimal modifications at the level of individual neurons or layers.

Recently, model editing methods have evolved into five main types based on how to adjust LLM's parameters with reference to the collected knowledge texts: 
(1) locate-then-edit methods \cite{meng2022locating, mengmass, li2024pmet} typically proceed in two stages: first, identifying specific neurons associated with the target knowledge during the \textit{locating} stage, and second, adjusting parameters within these locations to integrate the new knowledge during the \textit{editing} stage;
(2) meta-learning based methods \cite{de2021editing, tanmassive} aim to train an editor to \textit{learn to edit}, enabling adaptation to various knowledge updates;
(3) fine-tuning methods \cite{huangtransformer, yu2024melo, wang2024wise} often employ additional adapter layers or other auxiliary parameters to introduce new knowledge while preserving the integrity of the LLM's original parameters;
(4) external memory-based methods \cite{hartvigsen2024aging,mitchell2022memory} store updated knowledge in the form of embeddings or hidden states, or even the parameters of a smaller model; 
(5) representation editing methods \cite{hernandez2023inspecting, wu2024reft} dynamically adjust representations or hidden states during the generation process to incorporate new knowledge.

Although these knowledge editing methods have shown some successes, there are also studies that discuss their shortcomings, some of which are supported by evaluations, to be detailed in Section~\ref{sec:related}.  Modifying the parameters of an LLM for the sake of the limited knowledge updates can potentially cause more harm than good to its overall performance. Theoretically, model editing builds on insights from the interpretability of the LLM \cite{wang2024knowledge}. On the one hand, LLMs are assumed as knowledge bases, with the knowledge primarily stored in feed-forward network (FFN) layers as key-value repositories, while the attention modules facilitate the information flow \cite{zhumodifying, geva2021transformer}. On the other hand, it is assumed that knowledge in LLMs is distributed across both FFN and attention layers, rather than being confined to a single layer type or neuron \cite{allen2024physics}. This conflict leads to a crisis in the model editing methods of modifying or adding parameters to replace updated knowledge. As shown in Fig. \ref{fig:model_intro} (b), intuitively, this kind of point-to-point editing can only ensure that the LLM can possess the knowledge injected this time. The price could be that the general knowledge and capabilities of the post-edit LLM may be severely damaged \cite{gu2024model,wang2024missing,halevy2024flex}. Coupled with the aforementioned challenges in simultaneously collecting a comprehensive set of knowledge data, the post-edit LLM may struggle to engage in further reasoning with the latest knowledge \cite{cohen2024evaluating}. 
In addition, these particular model editing methods are limited to single edit \cite{meng2022locating}, rather than allowing for multiple edits sequentially \cite{yao2023editing,hartvigsen2024aging}. For simplicity, we take updating one piece of knowledge at a time as an example. A single edit can successfully adjust parameters to achieve the desired knowledge. But, after multiple edits for different knowledge, the corresponding places of the LLM's parameters will be modified sequentially. As a result, the original knowledge of the post-edit LLM is continuously impaired, rendering it incapable of meeting anticipated performance standards.

Yet, a substantial gap remains in the comprehensive empirical validation of these issues. Ideally, an effective model editing method should enable seamless knowledge updates within an LLM. This requires ensuring that the post-edit LLM provides precise and accurate responses to corresponding queries derived from the updated knowledge texts. Moreover, it's essential to recognize that factual knowledge is interconnected within a broader knowledge graph as shown in the lower half of Fig. \ref{fig:model_intro} (a). Therefore, the post-edit LLM should be capable of addressing multi-hop inquiries related to the newly integrated knowledge \cite{cohen2024evaluating}. At the same time, the post-edit LLM must preserve existing world knowledge and reasoning capabilities independent of the updates. When asked queries irrelevant to new knowledge, the post-edit LLM should maintain the same proficiency as before the updates.

To this end, this study considers four critical dimensions~\cite{zhang2024comprehensive} as shown in Fig. \ref{fig:model_intro} (d): \textbf{reliability}, which refers to the expectation that the post-edit LLM is supposed to output the new target regarding the edited prompt; \textbf{generalization}, which indicates that the post-edit LLM should accurately respond to paraphrased prompts; \textbf{locality}, which highlights the requirement for the post-edit LLM to maintain its original responses when encountering queries irrelevant to the update; and \textbf{portability}, which emphasizes that the post-edit LLM should consider the downstream effects of the edited knowledge, accounting for alternative names, reversed relationships, and further reasoning. Reliability and locality are key indicators for ensuring the accuracy and stability of knowledge updating, while generalization and portability demonstrate the post-edit LLM's capacity to integrate and apply updated knowledge across diverse contexts.

We highlight that, in our evaluation, during the inference phase, we employ an autoregressive setting \cite{mccoy2023embers} instead of a teacher-forced setting \cite{mitchellfast, melocang2022locating,zhang2024comprehensive, wang2024wise} to generate the next token, thereby constructing the target incrementally. This can prevent from exaggerating model performance when recalling knowledge, thereby obtaining a fair and accurate performance evaluation that is more in line with practical applications. 

Ten model editing methods are evaluated on two counterfactual datasets: WikiData\(_\text{counterfact}\) \cite{cohen2024evaluating} and ZsRE \cite{levy2017zero}.
The number of edits ranges from 10 to 100 to cover the entire dataset. Experimental findings indicate that \textit{no model editing method can achieve outstanding performance across all dimensions} (Section \ref{sec:assessment}). Furthermore, most existing methods lack robustness; their performance sharply declines as the number of edits increases (Section \ref{sec:assessment2}).
In particular, parameter update-based methods, such as ROME \cite{melocang2022locating}, MEMIT \cite{mengmass}, and MEND \cite{mitchellfast}, are particularly susceptible to catastrophic forgetting due to parameter overwriting. Continuously modifying parameters at various locations can undermine the effectiveness of previous edits and gradually erode the world knowledge retained by LLMs. WISE \cite{wang2024wise} introduces additional parameters and adopts a parameter fusion mechanism to integrate these increasing parameters during sequential editing. However, the inherent challenge of managing an excessive number of parameters may eventually hinder overall performance. Importantly, WISE is restricted to token-level corrections, making it less suitable for autoregressive generation settings.
Memory-based methods, such as GRACE \cite{hartvigsen2024aging}, can achieve good results in terms of reliability and locality. However, they still face challenges related to generalization and portability, as it remains heavily reliant on rote memorization.
Converting knowledge into parameters or hidden states could be a redundant action.

In light of the unreliability and significant shortcomings in knowledge updating performance of model editing approach, we recommend against altering model parameters for minor knowledge updates. Instead, we should fully leverage the contextual reasoning capabilities known as \textit{In-context Learning} inherent in LLMs \cite{brown2020language,ouyang2022training,dong2022survey}.
As shown in Fig. \ref{fig:model_intro} (c), in addition to providing direct instructions to an LLM, supplementary background information and illustrative examples can also be incorporated into the prompt. The LLM's understanding of this context can modulate the response of existing knowledge to the instruction, or even overwrite existing knowledge \cite{shumailov2024ununlearning}. Therefore, we propose a straightforward yet effective method called \textit{Selective Contextual Reasoning} (SCR) to facilitate knowledge updating (Section \ref{sec:method}).

Similar to existing model editing methods, our proposed SCR is grounded in the utilization of collected textual knowledge data. However, rather than altering the model parameters, SCR focuses on strategically selecting specific pieces of knowledge to incorporate into the prompt as contextual information. To accomplish this, SCR first assesses whether an incoming query falls within the scope of the collected knowledge data. Initially, a retriever extracts the most relevant information from the knowledge data in response to the query. The LLM is then prompted to evaluate this information for its relevance to the query. If deemed relevant, both the retrieved knowledge text and the query are incorporated into the prompt for the LLM to generate a response. Conversely, if the query is determined to be irrelevant, only the query is included in the prompt for the LLM to formulate a response.

We conduct extensive experiments to evaluate SCR against the aforementioned model editing methods (Section \ref{sec:assessment2}). We utilize three mainstream LLMs, including Llama-2-7B-chat \cite{touvron2023llama}, Llama-3.1-8B-instruct \cite{llama3modelcard}, and Mistral-7B-instruct \cite{jiang2023mistral}, as backbone LLMs. Experimental results indicate that SCR necessitates merely a straightforward two-step filtration of the most pertinent knowledge, devoid of any training, yet successfully attains a balanced performance across four dimensions.
The results indicate that updating the knowledge of an LLM doesn't always necessitate altering its parameters. Instead, selective contextual reasoning can yield excellent performance.

From a data collection perspective, we argue that SCR is more practical than the existing model editing models.  Because knowledge update happens along time, the knowledge pieces to be updated  cannot be as comprehensively collected as the pre-training data, as illustrated in Fig. \ref{fig:model_intro} (a). In practical settings, it is essential to \textit{continually} identify new knowledge derived from the evolving real world, which distinctly differ from the original knowledge retained by LLMs. When a knowledge update is confirmed, it can also be challenging to incorporate related updates comprehensively during the data collection phase. For example, if the knowledge update states that \textit{The President of the United States is Donald Trump}, it may be difficult to include additional context, such as details about the school that President Trump's grandchild attends. 

Our main contributions can be summarized as follows. First, we provide a comprehensive evaluation of ten existing model editing methods across four dimensions, under the more challenging yet realistic setting of autoregressive generation. The results indicate that most existing methods prove ineffective and ultimately counterproductive. Second, we propose a simple and effective approach, called selective contextual reasoning (SCR), as an alternative to traditional model editing methods for the continual knowledge update of LLMs. Third, we perform sequential knowledge updating based on the WikiData\(_\text{counterfact}\) and ZsRE datasets for Llama-2, Llama-3.1, and Mistral, and the experimental results demonstrate the effectiveness of our method SCR.

%================================
\section{Related Work}
\label{sec:related}
%================================

This paper is dedicated to revisiting knowledge updating in LLMs. We begin by reviewing existing model editing methods, followed by a critical analysis of concerns associated with these methods. Lastly, we review the emergence and applications of contextual reasoning capabilities within LLMs.

%=================================
\subsection{Model Editing Methods}
%=================================

\begin{figure} 
	\centering
	\includegraphics[width=0.7\textwidth]{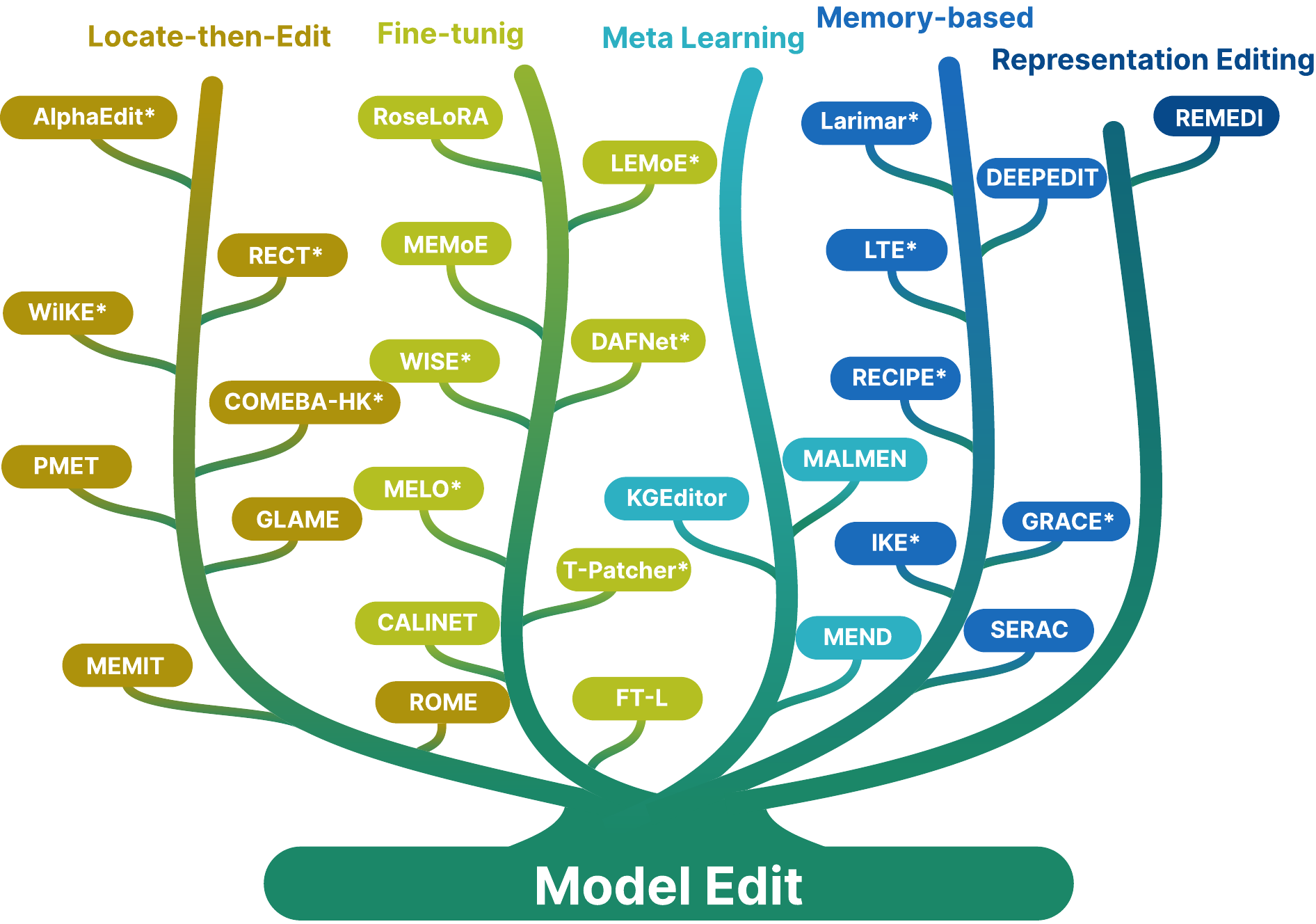}
        \caption{The five types of \textit{Model Editing} methods. Methods marked with an asterisk (*) are claimed to be suitable for sequential editing.}
	\label{fig:technology_tree}
\end{figure}

LLMs are often regarded as knowledge bases, as they encapsulate vast amounts of world knowledge within their extensive parameters, which are derived from large-scale datasets during the pre-training phase~\cite{petroni2019language, geva2021transformer, geva2023dissecting, dai2022knowledge}. 
To cope with knowledge updates, the model editing approach~\cite{meng2022locating, mengmass} encodes target knowledge into specific parameters, which are then replaced or supplemented in the LLM to update its factual knowledge. Existing model editing methods can be broadly categorized into five types, as illustrated in Fig.~\ref{fig:technology_tree}.

The \textbf{locate-then-edit} methods~\cite{mengmass, zhang2024knowledge, li2024pmet, li2024consecutive,hu2024wilke} are ground in the interpretability theory of Transformer architecture~\cite{geva2021transformer,lv2024interpreting}. It posits that knowledge is distributed across feed-forward networks (FFNs), while attention modules play a role in information copying and transmission. For instance, ROME~\cite{meng2022locating} utilizes causal tracing to first identify the crucial neurons associated with specific knowledge before performing targeted edits. 
Furthermore, RECT~\cite{gu2024model} and AlphaEdit~\cite{fang2024alphaedit} introduce additional constraints based on ROME to prevent excessive parameter shifts during the editing process.
Note that, the assumption of the localized storage of factual knowledge remains controversial~\cite{wei2024does}. 
An alternative hypothesis suggests that the relationship between neurons and knowledge is characterized by a many-to-many dynamic rather than a simplistic one-to-one association~\cite{allen2024physics}. Any modification to the parameters will inevitably affect other knowledge stored in the LLM, including both the original knowledge, and previously edited knowledge.

\textbf{Fine-tuning} methods aim to efficiently integrate target knowledge by isolating the parameters that require adjustment, such as WISE~\cite{wang2024wise}, T-Patcher~\cite{huangtransformer}, MELO~\cite{yu2024melo} and many others~\cite{dong2022calibrating, zhang2024dafnet, wang2024memoe, wang2024lemoe, wang2024roselora}. 
These methods typically introduce additional parameters or utilize mixture of experts (MoE) architectures~\cite{chen2022towards}, either at the head of the LLM or within its structure.
However, as the number of additional parameters increases, the likelihood of overfitting escalates. This phenomenon can result in the post-edit LLM neglecting prior edits, and compromising its original knowledge. Besides, the continual expansion of neurons may further exacerbate the post-edit LLM's inference burden.

\textbf{Meta-learning} methods, such as MEND~\cite{de2021editing} and MALMEN~\cite{tanmassive}, employ hyper-networks that are designed to forecast tailored weight updates for each knowledge data instance associated with an LLM. However, the hyper-network for a particular LLM limits their scalability in sequential editing scenarios. Furthermore, the additional training process incurs significant time and computational costs. Additionally, the necessity of modifying parameters for a limited amount of knowledge encapsulated in textual form is a matter of ongoing debate.

\textbf{Memory-based} methods~\cite{zhong2023mquake,hartvigsen2024aging,mitchell2022memory,jiang2024learning, chen2024lifelong, daslarimar} maintain a memory store for updated knowledge, which can be represented as plain text, hidden states, or token embeddings. SERAC~\cite{mitchell2022memory}, a classical method, simulates the editing scope by training a discriminator, whose results distinguish between the original LLM and the counterfactual model. GRACE~\cite{hartvigsen2024aging} maintains a dynamically updated codebook that alters the hidden states during the forward propagation.
Note that, these methods typically require frequent updates to the classifier as the memory expands, necessitating continuous training of the additional discriminator models.
In the deeper layers of large-scale neural models, most feature values diminish significantly, compressing the representational space into a limited set of feature directions. This phenomenon, known as dimensional collapse~\cite{dohmatob2024strong}, undermines the reliability of hidden states for encoding and retaining edited knowledge.

\textbf{Representation editing} methods, such as REMEDI~\cite{hernandez2023inspecting} and REFT~\cite{wu2024reft}, dynamically adjust the representations or hidden states during the generation process. They demonstrate the feasibility of control over the generation of LLMs. 
However, they also suffer from the problem of inflexible expansion of knowledge updating.

In short, most existing methods update LLM knowledge by modifying their parameters or structures. However, in real-world scenarios, the amount of new knowledge available is limited, compared to the vast pre-training data used by LLMs. Encoding new knowledge into the model’s parameters can lead to the loss of original knowledge, risking both an incomplete understanding of updates and potential conflicts with prior knowledge.

%==========================================
\subsection{Critiques of Model Editing}
%==========================================

Model editing allows for targeted modifications to an LLM's knowledge, but many studies suggest it may also introduce negative consequences. 
\citet{pinter2023emptying} criticize that direct model editing pursues factual accuracy, which is misaligned with the pre-training objectives of LLMs. They caution that model editing reinforces the flawed notion that \textit{model authenticity is reliable} and cannot serve as a remedy for the inherent shortcomings of LLMs. 
\citet{wang2024missing} quantitatively evaluates the negative impact of the ripple effect in the hidden space, which significant hinders the effectiveness of editing tasks and overall performance of post-edit LLMs. 
Similarly, \citet{gu2024model} demonstrates that improvements in factuality come at the cost of a significant decline in reasoning, natural language inference, and question-answering abilities. LLMs are not robust to parameter perturbations, as editing 1\% of parameters can cause a sharp decline in other tasks. Achieving both factual accuracy and general capability remains a challenging dual objective. 
\citet{yang2024butterfly} also experimentally confirms that after a few edits, the coherence of the post-edit LLM's text generation decreased significantly, leading its performance on downstream tasks to approximate random guessing. 
Additionally, \citet{yang2024fall} demonstrates that ROME could cause the LLM to crash with just a single edit. 
And \citet{halevy2024flex} finds that editing model parameters can exacerbate existing biases against certain demographic groups and amplify misinformation, intensifying issues like racial biases and gender discrimination to varying degree across all methods and models. 

These studies indicate that direct model editing is neither an effective nor a reliable solution for addressing knowledge obsolescence in LLMs. However, these critiques often focus on a single perspective, lacking a comprehensive evaluation of existing model editing methods. In this paper, we present a thorough assessment, examining all four dimensions of knowledge update goals within the context of autoregressive generation.

%=====================================
\subsection{Contextual Reasoning}
%=====================================

In-context Learning (ICL)~\cite{brown2020language} is a prominent emergent ability of LLMs~\cite{wei2022emergent}. 
ICL enables LLMs to learn and reason directly from contextual information, eliminating the need for explicit re-training.
By leveraging the patterns and structures acquired during pre-training, ICL uses examples or task-specific prompts within the context to help the LLM understand the task and generate appropriate outputs, which correspond to few-shot learning and zero-shot learning, respectively. 
By carefully organizing examples or designing instructions, more ICL techniques, such as Self Adaptive~\cite{wu2023self} and Self-instruct~\cite{wang2023self}, have been developed, further unlocking the potential of LLMs. 
When tackling complex tasks, Chain-of-Thought (CoT)~\cite{wei2022chain} uses magic prompts or introduces intermediate reasoning steps into the examples to help the LLM not only arrive at the answer but also understand the underlying reasoning process. 
LLMs perform conditional generation based on the given context, which actually exploits the capabilities of understanding and reasoning within the boundaries of its generalization abilities~\cite{li2023transformers}.

Constrained by the static nature of pre-training data, LLMs are limited in their awareness of recent events, and their intrinsic knowledge may be subject to inaccuracies, leading to hallucinations and ignorance~\cite{ji2023survey}. 
Retrieval-Augmented Generation (RAG)~\cite{lewis2020retrieval} seeks to bridge this gap by augmenting the LLM's intrinsic knowledge with real-time, relevant external information from knowledge bases or online sources. 
Through the integration of retrieval models and ICL, LLMs can effectively enhance their adaptability to new information while maintaining context-dependent language understanding. 

The key advantage of ICL is that there is no need for gradient backpropagation or re-training LLMs for specific tasks. Considering that fine-tuning may affect the LLM's general capabilities~\cite{luo2023empirical}, we explore a strategy that combines the selected knowledge text and contextual reasoning of LLMs to achieve the acquisition of new knowledge. 

%=================================
\section{Preliminaries}  
\label{sec:preliminaries}
%=================================

This paper explores the scenario of \textit{continuously} knowledge updating, \ie the objectives of sequential editing task, also known as lifelong editing or continual editing~\cite{hartvigsen2024aging,yu2024melo,wang2024wise}. We consider a piece of knowledge $e_i$ to be conceptualized as a triplet consisting of a subject, a relation, and an object. This triplet can be expressed either in a question-and-answer format or as a declarative statement. For easy presentation, we adopt the question-and-answer format, representing a piece of knowledge as $e_i = (x_i, y_i)$, where $x_i$ is the question and $y_i$ is the corresponding answer.
 
In our evaluation, we consider an LLM to be a mapping function $f$ which takes in an input question $x$ and produces its answer $y$. Given a sequence of knowledge pieces $E_t = \{ e_1, e_2, \ldots, e_t \}$ for updating an LLM, we aim to implement a continuous updating process. Starting from the original LLM, denoted by $f^0$, this process produces a series of post-edit LLMs, each $f^t$ reflecting the incorporation of the $t$-th knowledge piece.

After incorporating $t$ pieces of new knowledge, \ie undergoing $t$ rounds of updates, the post-edit LLM $f^t$ is evaluated in four dimensions~\cite{zhang2024comprehensive}:

\begin{itemize}
    \item \textbf{Reliability:} The post-edit LLM should generate the updated target output for the prompts used for knowledge editing, \ie those available in $E_t$,  and ensure the persistence of the editing effects. This can be formally expressed as:
    \begin{equation}
        \mathbb{E}_{(x_i, y_i) \sim E_t} \mathbb{I} \{ \arg\max_y f^t(y \mid x_i) = y_i \}.
    \end{equation}

    \item \textbf{Generalization:} The post-edit LLM should extend beyond the exact edits and correctly respond to paraphrased prompts, denoted as $N(E_t)$. Mathematically, this is formulated as:  
    \begin{equation}
        \mathbb{E}_{(x_i, y_i) \sim N(E_t)} \mathbb{I} \{ \arg\max_y f^t(y \mid x_i) = y_i \}.
    \end{equation}

    \item \textbf{Locality:} It is imperative that the post-edit LLM should retain its original behavior when processing queries that are unrelated to the edits, denoted by $O(E_t)$. This criterion measures how well the post-edit LLM maintains the overall stability while keeping edits confined to the relevant scope. This requirement can be represented as:
    \begin{equation}
        \mathbb{E}_{(x_i, y_i) \sim O(E_t)} \mathbb{I} \{ f^t(y \mid x_i) = f^0(y \mid x_i) \}.
    \end{equation}

    \item \textbf{Portability:} Finally, the post-edit LLM should effectively propagate the impact of the edited knowledge, correctly reasoning about its downstream implications, denoted by $D(E_t)$. $D(E_t)$ encompasses three aspects: substituting the subject of the question with aliases, reasoning based on factual changes, and knowledge derived from reverse relationships. The requirement is defined as:
    \begin{equation}
        \mathbb{E}_{(x_i, y_i) \sim D(E_t)} \mathbb{I} \{ \arg\max_y f^t(y \mid x_i) = y_i \}.
    \end{equation}
\end{itemize}

To summarize, we evaluate the responses of an edited LLM against: 
(i) $E_t$, the questions used for knowledge editing;  
(ii) $N(E_t)$, rephrased versions of the questions used for knowledge editing;  
(iii) $O(E_t)$, questions unrelated to those in $E_t$; and  
(iv) $D(E_t)$, questions reflecting downstream implications of knowledge editing, including subject alias substitution, reasoning based on factual changes, and knowledge derived from reverse relationships.  

If the edited LLM adheres to these requirements, then the updating process is considered robust, precise, and minimally disruptive to unrelated LLM behavior.

%======================================
\section{Assessment of Model Editing}
\label{sec:assessment}
%======================================

In this section, we conduct a thorough assessment of ten prominent model editing methods. We use counterfactual knowledge as the datasets for model editing, and evaluate their accuracy along the four aforementioned dimensions. Next, we detail the experiment setting. 

%======================
\subsection{Experiment Settings}
%======================

%======================
\subsubsection{Datasets}
%======================

To ensure a fair comparison among various LLMs with differing training cutoffs, we employ counterfactual knowledge to serve as updated knowledge datasets. This guarantees that the information to be updated does not appear in the original training sets of the LLMs.  

Specifically, we utilize two widely used context-free question-answering (QA) datasets, namely \textbf{WikiData\(_\text{counterfact}\)}~\cite{cohen2024evaluating} and Zero-Shot Relation Extraction (\textbf{ZsRE})~\cite{levy2017zero}. Both datasets are well-established counterfactual benchmarks designed for modifying the LLM's current knowledge. 
The statistics of these two datasets are presented in Table~\ref{tab:statistics}. Notably, in this study, all model editing methods, with the exception of MEND and WISE, do not necessitate a training set.

\begin{itemize}
\item \textbf{ZsRE}: Originally a question-answering dataset, ZsRE is extended by \citet{yao2023editing} to assess various dimensions of model editing methods. 

\item \textbf{WikiData\(_\text{counterfact}\)}: This dataset collects triplets about popular entities, ensuring that the subject corresponds to one of the most viewed Wikipedia pages. 
\end{itemize}

\begin{table}
{\small 
  \centering
  \caption{Statistics of Training and Testing Sets for Wiki$_\text{counterfact}$ and ZsRE Datasets.}
  
    \begin{tabular}{c|cc}
    \toprule
    \textbf{Dateset} & \textbf{Wiki$_\text{counterfact}$} & \textbf{ZsRE} \\
    \midrule
    \#Train & 1427  & 10000 \\
    \#Test & 839   & 1301 \\
    \bottomrule
    \end{tabular}
  \label{tab:statistics}
  }
\end{table}

Each entry in these datasets includes the edited question along with its corresponding target, as well as one or more question-target pairs designed to assess key attributes of the model edit method, such as reliability, portability, and generalization, as discussed in Section~\ref{sec:preliminaries}. 
To increase the challenge, the locality data includes questions that pertain to the same subject as the editing target but explore different relationships.  
Illustrative examples from these datasets are provided in Tables~\ref{tab:datasetExample}.

\begin{table}
  \centering
  \caption{Example questions from the two datasets, with different question types and targets.}
  \resizebox{\textwidth}{!}{
    \begin{tabular}{l|lll}
    \toprule
 &    \textbf{Type}  & \textbf{Question} & \textbf{Target} \\
    \midrule
    \multirow{6}{*}{\rotatebox[origin=c]{90}{Wiki$_\text{counterfact}$}}        
    & Reliability & The name of the country of citizenship of Leonardo DiCaprio is & Syria \\
     & Generalization & Leonardo DiCaprio's country of citizenship is known as & Syria \\
    & Locality$_\text{Relation Specificity}$ & The name of the mother of Leonardo DiCaprio is & Irmelin DiCaprio \\
    & Locality$_\text{Forgetfulness}$ & The name of the country of citizenship of Leonardo DiCaprio, which is not Syria, is &  America \\
    & Portability$_\text{Subject\_Aliasing}$ & The name of the country of citizenship of Di Caprio is & USA \\
    & Portability$_\text{Reasoning}$ & The name of the currency in the country of citizenship of Leonardo DiCaprio is & Syrian pound \\
    \midrule
    \multirow{4}{*}{\rotatebox[origin=c]{90}{ZsRE}}
    & Reliability & Which family does Epaspidoceras belong to? & Noctuidae \\
    & Generalization & What family are Epaspidoceras? & Noctuidae \\
    & Locality$_\text{Relation Specificity}$& The taxon rank of Epaspidoceras is & genus \\
    & Portability$_\text{Reasoning}$ & What is the common name for the family Epaspidoceras belongs to? & Owlet moths \\
    \bottomrule
    \end{tabular}
  }
  \label{tab:datasetExample}
\end{table}

%=============================
\subsubsection{Backbone LLMs}
%=============================
For the evaluation experiment presented in this section, we use Llama-2-7b-chat \cite{touvron2023llama} as the base LLM. As a representative of decoder-only architectures, the LLaMA model family delivers impressive performance. 

%=============================
\subsubsection{Target of Editing}
%=============================
The experimental scenario centers on sequential editing \cite{zhang2024comprehensive}, where modifications are made one at a time, and the results are evaluated across the entire sets of edits once all changes are applied.
In this session, each piece of knowledge across the entire dataset is updated individually, and the evaluation is conducted after all updates have been conducted.

%=============================
\subsubsection{Inference Strategy}
%=============================
Following methods like MEND \cite{mitchellfast} and ROME \cite{melocang2022locating}, several subsequent studies \cite{zhang2024comprehensive, wang2024wise} have adopted teacher forcing \cite{williams1989learning} during the inference phase, resulting in a certain degree of artificially inflated performance that may not entirely and accurately reflect the true capabilities of these methods. While some researchers \cite{wang2024wise} argue that changes in token prediction under teacher forcing indicate a successful influence on the LLM, this approach is considered unrealistic for real-word predictions and, to some extent, can be seen as a form of cheating. Therefore, for a more equitable and reasonable assessment, we uniformly adopt an \textbf{autoregressive generation} \cite{mccoy2023embers} paradigm for prediction.

%=============================
\subsubsection{Metrics}
%=============================

We use \textit{accuracy} as the metric to assess the methods following the four evaluation dimensions~\cite{zhang2024comprehensive} introduced in Section \ref{sec:preliminaries}: reliability, generalization, locality, and portability.
\begin{equation}
\text{Accuracy} = \frac{1}{T} \sum_{t=1}^{T} \mathbb{I}\left(y_t = \hat{y}_t\right).
\end{equation}
In the above formula:
\begin{itemize}
    \item \( T \) denotes the total number of prompts evaluated.
    \item \( y_t \) represents the target tokens for the \( t \)-th prompt.
    \item \( \hat{y}_t \) is the predicted tokens generated by the post-edit LLM for the \( t \)-th prompt.
    \item The indicator function \( \mathbb{I} \) equals \( 1 \) if \( y_t = \hat{y}_t \) (indicating an exact match) and \( 0 \) otherwise.
\end{itemize}

Note that, many previous studies have focused on token-level accuracy, with values ranging between 0 and 1. However, the goal of evaluation is to assess the post-edit LLM's mastery or forgetting of specific knowledge. A minor difference in a single token can lead to substantial variations in both individual words (which often consist of multiple tokens) and entire answers (which are typically composed of multiple words). 
Given the generative nature of LLMs, achieving a strict exact match with the target answer is often challenging. Therefore, as in many question-answering experiment setups \cite{asaiself}, we use a relaxed version of exact match (EM) by simply checking \textit{if the standard answer appears within the first 30 tokens} of the post-edit LLM's output, rather than requiring a strict exact match. For locality, we compare the accuracy before and after editing to measure how much of the original correct knowledge has been retained. Further, in our discussion, we report accuracy as a percentage (0 to 100) without using the \% sign.

\subsubsection{Baselines}
We utilize ten model editing methods as baselines, covering major categories of model editing methods. 

\textbf{ROME} \cite{melocang2022locating} identifies key neuron activations in factual predictions through causal interventions and updates specific facts by modifying the weights of the feed-forward network (FFN) in the intermediate layer.
\textbf{MEMIT} \cite{mengmass} specializes in batch editing, enabling simultaneous modifications of multiple facts. It extends the ROME framework by applying updates across multiple layers. 
\textbf{PMET} \cite{li2024pmet} takes into account that existing methods focus primarily on the Feed-Forward Network (FFN) module. It recognizes that the information flow comes from three components: Multi-Head Self-Attention (MHSA), Feed-Forward Network (FFN), and residual connections. Therefore, it simultaneously optimizes both MHSA and FFN to facilitate more effective knowledge updates. 
\textbf{RECT} \cite{gu2024model} is a regularization method that constrains weight updates based on ROME. Specifically, RECT preserves the original values of the top-$k$\% of edited weights, capturing the primary editing information. The remaining elements, viewed as having a secondary contribution, are zeroed out during regularization. 
\textbf{AlphaEdit} \cite{fang2024alphaedit} mitigates knowledge disruption in LLMs by projecting perturbations onto the null space of preserved knowledge before applying them, ensuring that original knowledge remains unaffected. 
\textbf{MEND} \cite{mitchellfast} trains a lightweight hypernetwork to convert gradients into weight updates while significantly reducing the number of parameters through low-rank decomposition. 
Since this work employs autoregressive generation for predictions, the locality outputs in MEND's training data are produced by the model itself, rather than being directly derived from the dataset. 
\textbf{FT-L}, proposed as a baseline of \cite{melocang2022locating}, directly fine-tunes a single Feedforward Neural Network layer. 
\textbf{AdaLoRA} \cite{zhang2023adaptive} is a parameter-efficient fine-tuning (PEFT) method that improves upon LoRA \cite{hulora} by adaptively adjusting module ranks. 
\textbf{GRACE} \cite{hartvigsen2024aging} introduces a discrete key-value module for storing edited hidden states without altering the weights, and it employs a deferral mechanism to decide whether to utilize the codebook. 
\textbf{WISE} \cite{wang2024wise} utilizes side memory to store edited knowledge, enabling collaboration between long-term and working memory. It maintains optimal knowledge density by sequentially storing batches of edits and merging parameters. Additionally, a router is trained to direct information to either the side memory or the main memory.

%=============================
\subsection{Results}
%=============================

\subsubsection{Characteristics and Efficiency}
\begin{table}
  \centering
  \caption{Characteristics and efficiency  of the ten model editing methods. \textit{Sequential Edit} indicates whether the method supports the sequential editing scenario. \textit{Edit Time} refers to the number of seconds required to edit a single fact. \textit{Inference Time} refers to the number of seconds required to edit a single fact. \textit{Training-free} indicates whether the editor can function without requiring training. \textit{Parameter-free} indicates whether the model's parameters remain unchanged—meaning no modifications or additional parameters are introduced. \textit{Interpretability} means whether the inference of the post-edited model is interpretable. \textit{Additional Data} refers to whether additional, edit-independent data is required during the editing or training process.}
  \resizebox{\textwidth}{!}{
    \begin{tabular}{c|crc|cccc}
    \toprule
    \textbf{Method} & \textbf{Sequential Edit} & \textbf{Edit Time} & \textbf{Inference Time} & \textbf{Training-free} & \textbf{Parameter-free} & \textbf{Interpretability} & \textbf{Additional Data} \\
    \midrule
    ROME  & \xmark     & 12.0153 & 1.6435 & \cmark    & \xmark    & \xmark    & \cmark  \\
    MEMIT & \xmark    & 9.7272 & 1.6567 & \cmark    & \xmark    & \xmark    & \cmark  \\
    PMET  & \xmark    & 11.3835 & 1.6697 & \cmark    & \xmark    & \xmark    & \cmark  \\
    RECT  & \cmark    & 13.6461 & 1.6929 & \cmark    & \xmark    & \xmark    & \cmark  \\
    AlphaEdit & \cmark    & 16.0965 & 1.7819 & \cmark    & \xmark    & \xmark    & \cmark  \\
    MEND  & \xmark    & 0.2796 & 1.5568 & \xmark    & \xmark    & \xmark    & \cmark  \\
    FT-L  & \xmark    & 2.6798 & 1.6734 & \cmark    & \xmark    & \xmark    & \xmark \\
    AdaLoRA & \xmark    & 19.4341 & 2.2005 & \cmark    & \xmark    & \xmark    & \xmark \\
    GRACE & \cmark    & 4.3308 & 1.6502 & \cmark    & \cmark    & \xmark    & \xmark \\
    WISE  & \cmark    & 21.1632 & 1.6957 & \cmark    & \xmark    & \xmark    & \cmark  \\
    \bottomrule
    \end{tabular}
  }
  \label{tab:existing method characteristics}
\end{table}

First, we analyze the characteristics and efficiency of evaluated model editing methods, as summarized in Table \ref{tab:existing method characteristics}. 

Methods such as RECT, AlphaEdit, GRACE, and WISE claim to support sequential knowledge updating. In contrast, approaches like ROME, MEMIT, PMET, MEND, FT-L, and AdaLoRA are designed for single or batch editing, which constrains their applicability in dynamic environments requiring incremental updates.
Several methods, including MEND, GRACE, and FT-L exhibit relatively low edit times, indicating high efficiency. However, MEND requires additional training, which incurs extra time and computational costs. 
GRACE stands out as the only approach with a parameter-free property, maintaining the original model parameters and avoiding the introduction of new ones. Most model editing methods, by contrast, rely on parameter or structure adjustments to implement knowledge updating. 
A significant limitation across these approaches is the lack of interpretability in the inference process. Furthermore, many approaches rely on additional training data to define boundaries for edits, which introduces drawbacks such as increased overhead for data collection and restricted knowledge generalization.

Overall, existing methods are constrained in their scenarios and fail to decouple fully from model parameters, falling short of optimal performance in both editing efficiency and interpretability, leaving room for further advancements in the field.

\subsubsection{Effectiveness}
\begin{figure}
    \centering
    \includegraphics[width=0.8\textwidth]{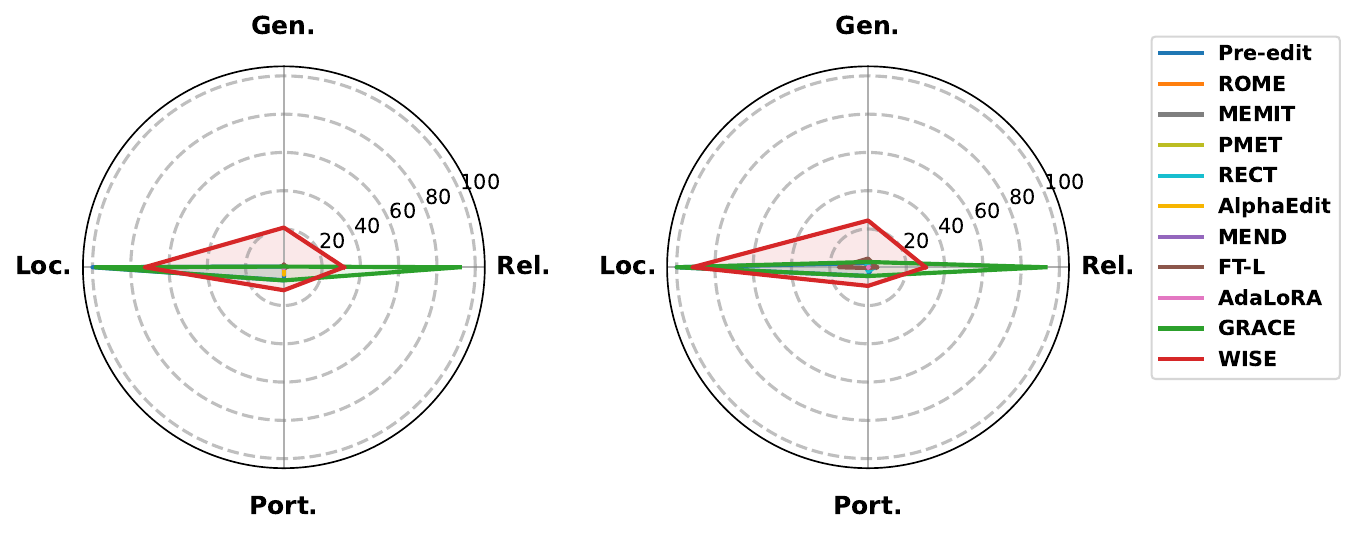}
   
    \caption{Accuracy of the ten methods on four dimensions: \textbf{Rel}iability, \textbf{Gen}eralization, \textbf{Loc}ality, and \textbf{Por}tability. The results are obtained by using Llama 2 as the base model, with autoregressive generation driven by model predictions. \textit{Left}: WikiData\(_\text{counterfact}\); \textit{Right}: ZsRE.}
    \label{fig:llama2_result}  
\end{figure}

Next, we assess the ten model editing methods across four key dimensions, namely reliability, generalization, locality, and portability.
The performance of editing Llama-2 on WikiData\(_\text{counterfact}\) and ZsRE is reported in Fig. \ref{fig:llama2_result}. 

\textbf{Reliability.} 
The sequential editing success rates of methods like ROME, MEMIT, MEND and PMET, which modify model parameters, are close to zero. Among all methods, GRACE achieves a reliability of over 90.0 on both datasets. Overall, most methods \textit{struggle to retain prior edits after incorporating subsequent updates}, resulting in consistently low reliability in model performance. 
Many approaches attempt to update specific knowledge by implementing fine-grained control. These operations typically involve local updates that alter a specific subset of model parameters or hidden states.
However, this strategy can inadvertently lead to the overwriting of critical parameters.
Even WISE, which employs a parameter merging mechanism, still suffers from forgetting issues within its side memory.
Therefore, given the assumption of factual knowledge locality remains unverified, the effectiveness of model edit methods based on this premise is questionable, as confirmed by the experimental results.

\textbf{Generalizability.}
There is considerable variation in the generalization performance across different methods. On WikiData\(_\text{counterfact}\) and ZsRE, WISE leads with relatively high scores of 20.74 and 24.44, respectively. In contrast, other methods, such as MEMIT, PMET, and MEND, show a complete lack of generalization.
The generalization metric requires the post-edit LLM to not only provide correct responses to the edited query but also to accurately handle rewritten or paraphrased versions of the same question. However, while methods like GRACE have demonstrated high edit success rates, they fall short in generalizing across different representations of the same knowledge. This limitation stems from the fact that their edits are restricted to adjusting token distributions, without equipping the LLM to truly understand the underlying knowledge.
These findings highlight the need for effective generalization strategies that beyond surface-level edits. A robust approach should empower the LLM to internalize and establish conceptual connections between various representations of knowledge in diverse contexts.

\textbf{Locality.}
GRACE demonstrates exceptional specificity, achieving perfect or near-perfect performance on both datasets by employing a radius-based mechanism to intercept hidden states that are inconsistent with the edits. WISE follows closely, reaching up to 91.80 specificity on ZsRE and 72.72 on WikiData\(_\text{counterfact}\). In contrast, ROME, MEMIT, and MEND exhibit significant forgetting of prior knowledge.
The challenges posed by different datasets differ considerably. While WISE performs well in ZsRE based on semantics, its performance falters in the WikiData\(_\text{counterfact}\) dataset, where prompts often share semantic similarities but reference distinct pieces of knowledge.
These findings indicate that even a finely tuned adjustment to a single neuron parameter can inadvertently influence global knowledge, highlighting the intricacies of maintaining locality during model editing.

\textbf{Portability.}
Among the evaluated methods, only WISE demonstrates some degree of portability, achieving a score of approximately 10.0. However, its impact remains minimal. Other mainstream methods largely neglect the ripple effect of knowledge propagation.
The primary reason existing methods struggle with portability is that they rely heavily on rote memorization of tokens, rather than the deep integration of knowledge. While model editing allows the LLM to memorize discrete pieces of knowledge, it does not lead to genuine comprehension of the knowledge. As a result, when the model encounters situations where similar knowledge needs to be expanded, transferred, or applied flexibly, existing methods often fall short.

In conclusion, portability remains a significant challenge for most current knowledge editing methods. In cases where research on knowledge association and interpretability within LLMs is still underdeveloped, explicitly presenting knowledge has emerged as an effective solution. By making knowledge explicit, the LLM no longer relies solely on implicit parameters or states adjustments to remember information. Instead, it can directly access clear, defined knowledge, making it easier to transfer and apply existing knowledge across different contexts.

\subsection{Failure Analysis}
We conduct a failure analysis using ZsRE dataset, focusing on four representative methods from the primary categories of model editing methods.

\begin{figure}
    \centering
    \includegraphics[width=0.4\textwidth]{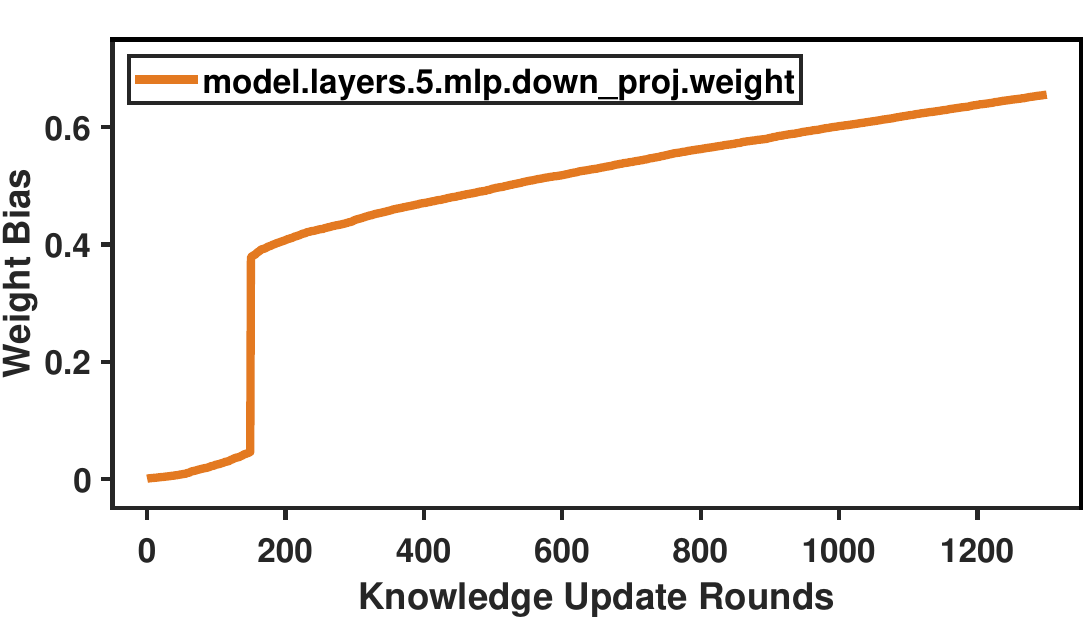}
    \hspace{0.01\textwidth}  
    \includegraphics[width=0.395\textwidth]{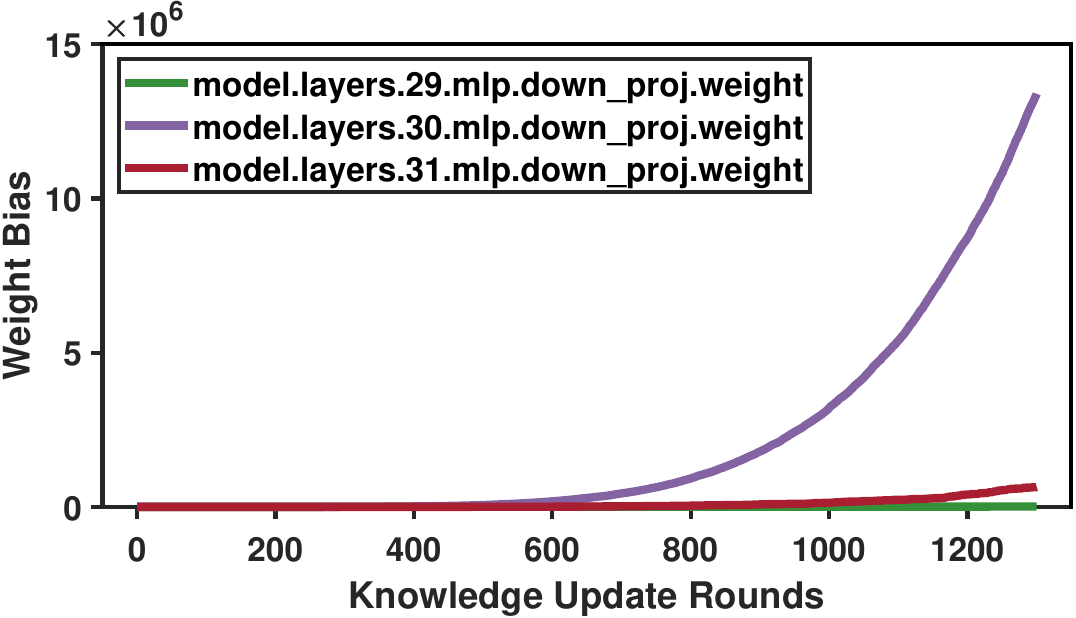}
    \caption{The weight bias of the edited layers gradually accumulate as the number of edits increases. \textit{Left}: ROME; \textit{Right}: MEND.}
    \label{fig:weight_bias}  
\end{figure}

For the locate-then-edit method ROME, and the meta learning-based method MEND, we visualize the parameter bias during sequential editing, as shown in Fig. \ref{fig:weight_bias}. The absolute values of parameter changes accumulate progressively, deviating significantly from the initial parameters and leading to catastrophic forgetting of previously updated knowledge. This highlights the inherent limitations of localized parameter updates, which prove ineffective in the context of sequential editing.

\begin{figure}
    \centering
    \includegraphics[width=0.5\textwidth]{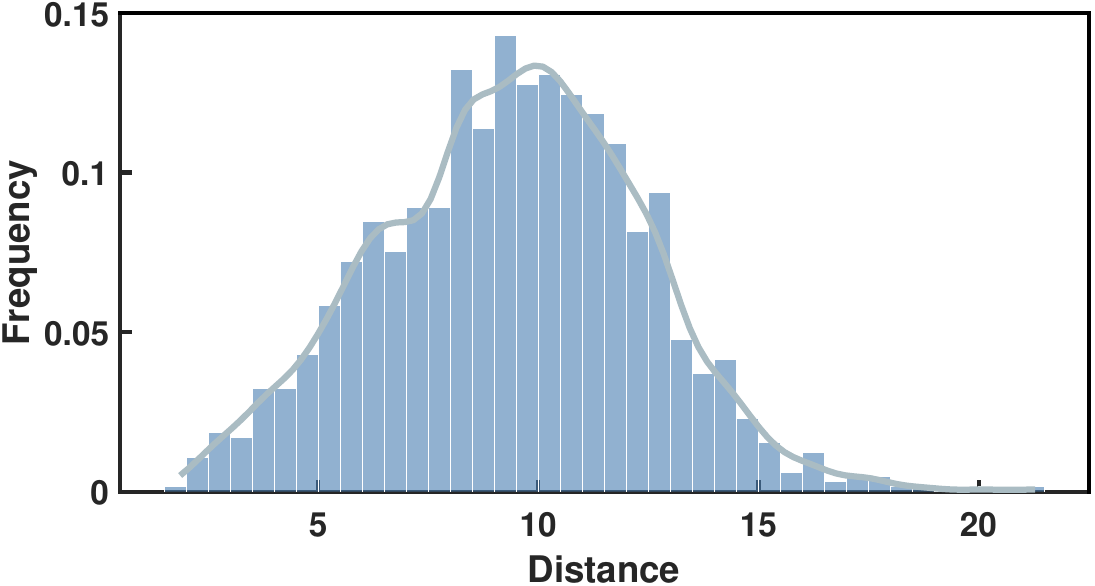}

    \caption{Distances between queries in rephrased prompts and codebook keys in the GRACE method.}
    \label{fig:GRACE_dist}  
\end{figure}

\begin{figure}
    \centering
    \includegraphics[width=0.5\textwidth]{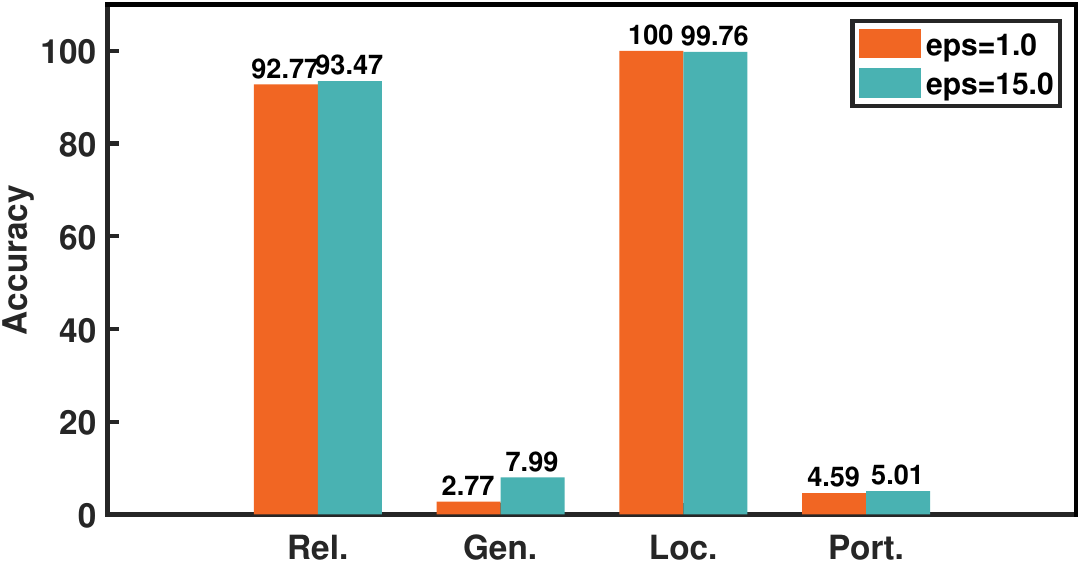}

    \caption{Performance on ZsRE as the radius of the deferral mechanism of GRACE increases from 1.0 to 15.0.}
    \label{fig:GRACE_eps}  
\end{figure}

Fig. \ref{fig:GRACE_dist} illustrates the distance between the hidden states of the rephrase prompts and their corresponding queries stored in the codebook for GRACE. The results suggest that the threshold hyperparameter defined in the original paper is inadequate for capturing the knowledge relevance between rephrased and original prompts. Even when the threshold radius is expanded to 15.0, as shown in Fig. \ref{fig:GRACE_eps}, the observed improvement in generalization remains marginal. The representation of a single token's hidden state in the codebook lacks sufficient expressiveness, rendering it ineffective for clustering semantically related prompts.

\begin{figure}
    \centering
    \includegraphics[width=0.3\textwidth]{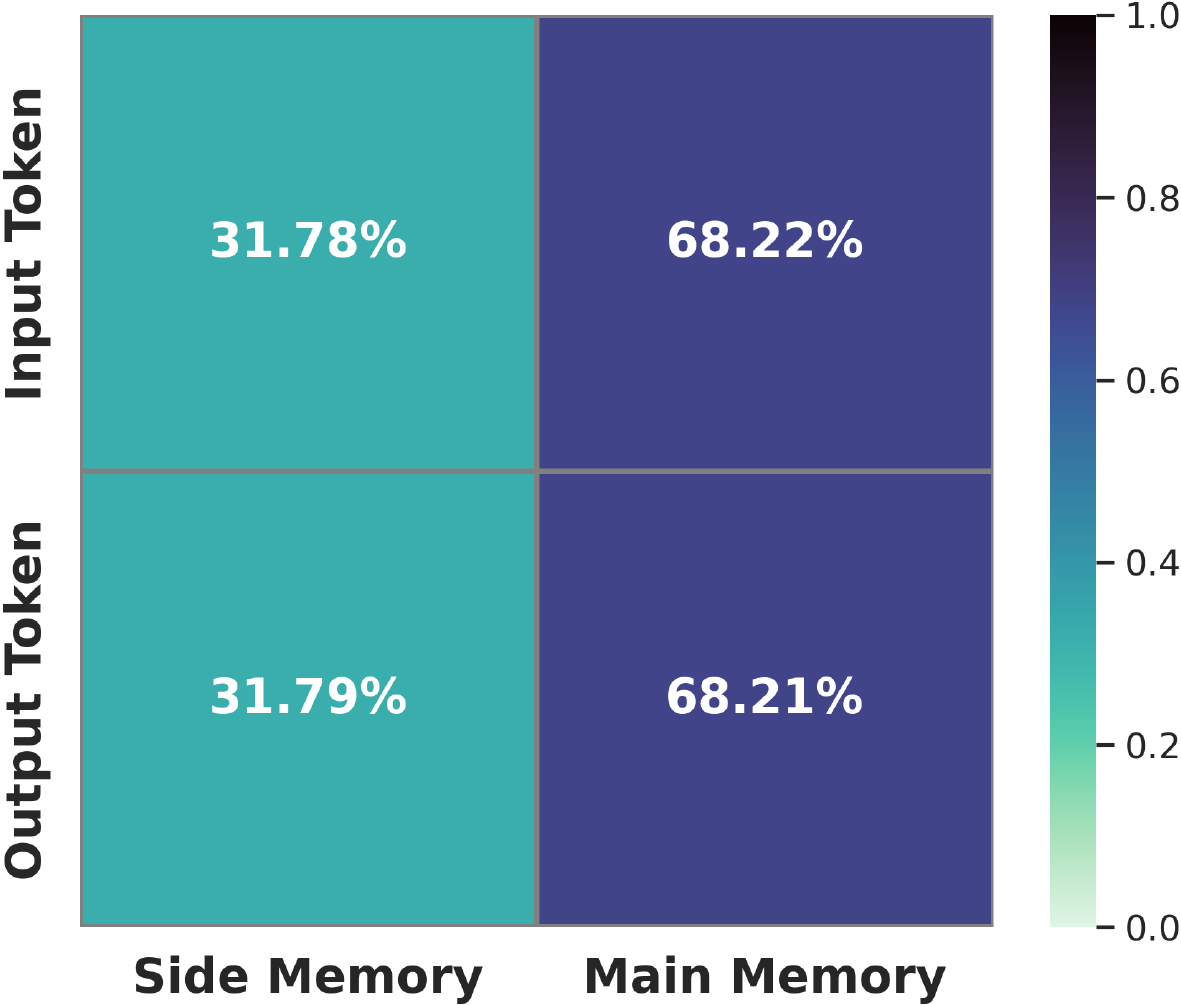}

    \caption{Proportions of input and output tokens allocated to side vs. main memory in WISE method for editing Llama-2 using ZsRE with the enhanced retrieval mode proposed in WISE.}
    \label{fig:wise} 
\end{figure}

For WISE, we adopt the retrieval mode mentioned in the original paper, which demonstrates superior performance compared to the merge mode.
Fig. \ref{fig:wise} shows the distribution of input and output tokens processed by the side memory and main memory, respectively. The results reveal that only about 30\% of tokens are handled by the correct memory, revealing the instability and unreliability of the method.

\noindent 

\subsection{Summary}
In the autoregressive generation setting, existing model editing methods either fail to perform well on any single metric or sacrifice one aspect to improve another, ultimately failing to achieve a balanced performance across all four dimensions. This underscores the need for a more holistic approach to knowledge editing. Moreover, these methods struggle to optimize both effectiveness and efficiency simultaneously, exposing a critical gap in current methodologies.

%=================================
\section{Knowledge Updates via Selective
Contextual Reasoning}
\label{sec:method}
%=================================
The extensive evaluation of existing methods suggests that modifying the parameters of an LLM for knowledge updates can potentially do more harm than good to its overall performance. Instead, knowledge updates should be handled by providing external contexts while keeping the LLM's parameters frozen. 

%=================================
\subsection{Overview}
%=================================

We propose a \textbf{Selective Contextual Reasoning (SCR)} framework that leverages an expandable memory for knowledge updates, relying on the reasoning and comprehension capabilities of LLMs.
SCR keeps the LLM's parameters frozen and dynamically updates its knowledge by selecting and contextualizing external knowledge memory based on the current query. In our implementation, the knowledge memory is a textual knowledge base. 

\begin{figure*}
	\centering
	\includegraphics[width=0.9\textwidth]{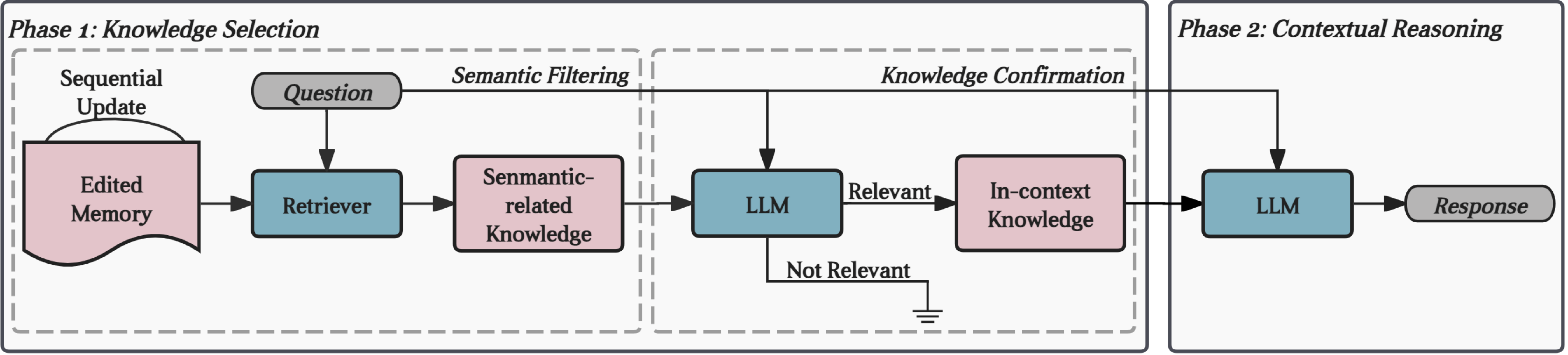}
	\caption{The Selective Contextual Reasoning (SCR) framework. The \textit{Edited Memory} is a dynamic textual knowledge base that can be expanded as needed. \textbf{Phase 1:} The retriever first applies semantic filtering to gather relevant information from memory based on the input question. The LLM then performs knowledge confirmation, assessing the alignment between the question and the retrieved knowledge. \textbf{Phase 2:} The LLM conducts conditional generation using the in-context knowledge, and the query.}
	\label{fig:mymode_method}
\end{figure*}

As illustrated in Figure~\ref{fig:mymode_method}, the core components of SCR are two-step knowledge selection and contextual reasoning. In the \textit{knowledge selection}, a retrieval model screens knowledge candidates from an external text collection based on semantic similarity to the query.
Then, the LLM evaluates these candidates for relevance, selecting the most pertinent knowledge text.
\textit{Contextual reasoning} enables the LLM to perform conditional generation by synthesizing instructions, updated knowledge, and queries to produce accurate responses.

%==================================
\subsection{Knowledge Selection}
%==================================

%==================================
\subsubsection{Semantic Filtering}
%==================================
Text serves as the vessel for knowledge, preserving information in its raw and interpretable state, unlike model parameters that primarily encode features and patterns. Thus, the memory, conceptualized as a dynamic textual knowledge base, is designed to expand continuously as new information becomes available, thereby facilitating the needs of sequential editing. 

Even though current LLMs have the capacity to handle long contexts, the rapid iteration of knowledge will eventually exceed the LLM's processing capacity. Further, not all texts are relevant to the current query. Therefore, we rely on a retriever to identify relevant knowledge based on the semantic similarity between the query and stored facts.
The selection can be based on any measures and in our implementation, we use cosine similarity between the query and the factual statements, and select top-$k$ ($k\geq 1$) retrieved  statements as candidates. 

The expandable memory allows for continuous updates without the risk of information loss due to parameter modification. And retrieval offers interpretability by directly tracing responses back to their textual sources, enabling easier validation and trustworthiness.

\subsubsection{Knowledge Confirmation}
While general retrieval offers information pertinent to a query, it often fails to deliver insights that expand upon the retrieved content.  Moreover, although similarity measure potentially help to filter out irrelevant knowledge texts, they struggle with highly deceptive queries where the subject aligns with the updated knowledge but the relationship differs. Thus, semantic relevance may not inherently guarantee the relevance of knowledge. 
This highlights the necessity for further meticulous selection. 

In this phase, the retrieved statements are evaluated and refined by the LLM to ensure its relevance and alignment with the query.
This process guarantees that the knowledge is both accurate and applicable to the specific query. 

Let $R_k(q)$ be the top-$k$ candidate factual statements selected from the external memory that are relevant to the query $q$. Depending on the setting of $k$, if only one statement is to be retrieved, \ie $k=1$, we will ask the LLM to judge whether the retrieved statement is indeed relevant to the query $q$. The statement will be used as external knowledge only if the LLM responds positively. More commonly, $k > 1$, we provide all the retrieved statements along with the query to the LLM and let the LLM select the most relevant statement. The selected statement will then be used as external knowledge. If none of the retrieved statements are deemed relevant to the query by LLM, we proceed without external knowledge.

\begin{figure}
	\centering
	\includegraphics[width=0.45\textwidth]{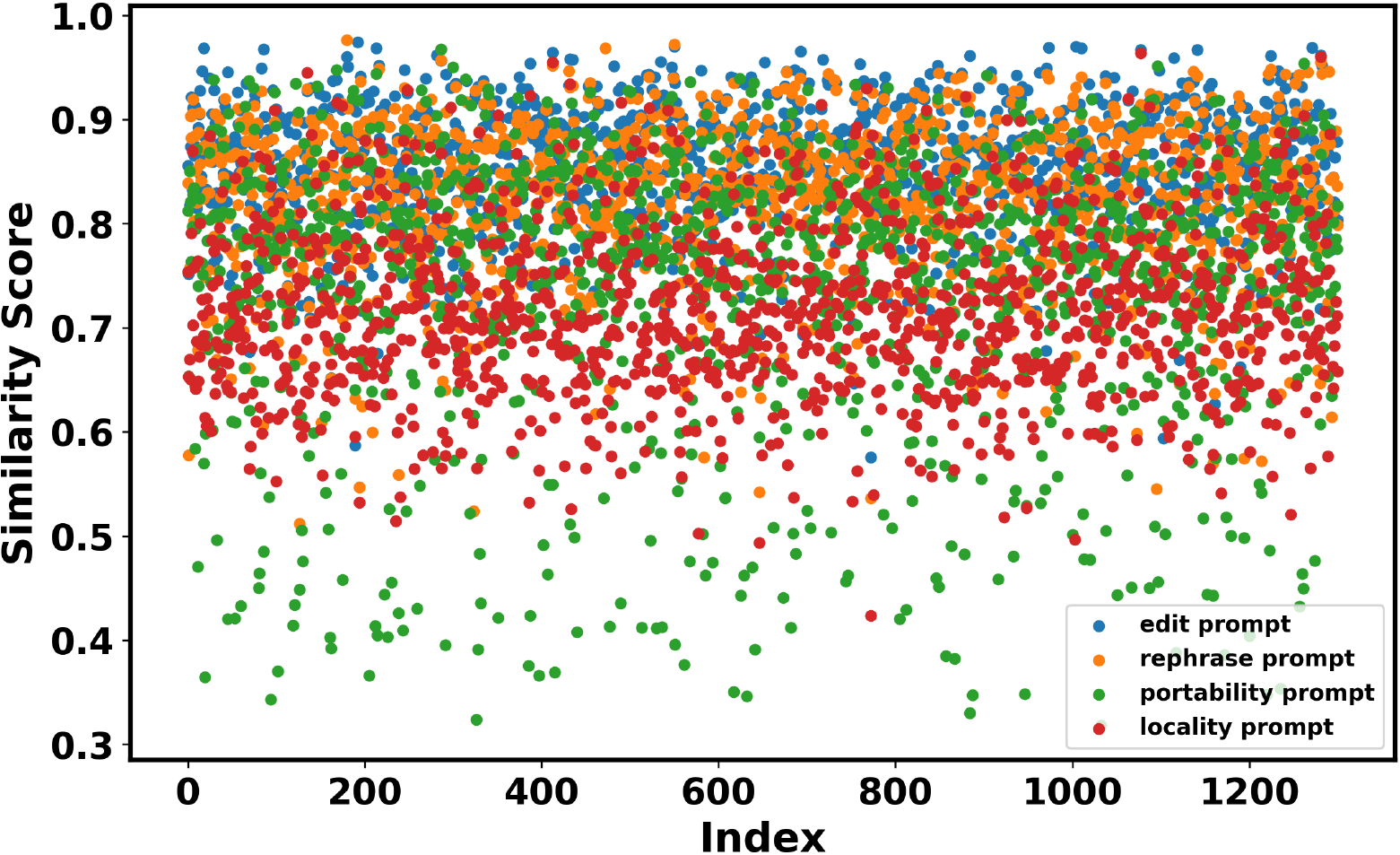}
	\caption{The top-\textit{1} similarity score between the four types of prompts and memory in the dataset.}
	
	\label{fig:similarity_score}
\end{figure}

To explore the importance of using LLMs for confirmation, we visualize the similarity score between four types of prompts and memory. As shown in the Fig. \ref{fig:similarity_score}, the locality prompt focuses on different attributes of the same subject, making it highly misleading and exhibiting a high degree of similarity with memory. Therefore, relying solely on similarity scores is insufficient to distinguish the scope of knowledge updating.

%==============================
\subsection{Contextual Reasoning}
%==============================

At this stage, the most relevant and up-to-date knowledge has  been identified and selected. 
SCR now integrates this knowledge with the original query to form a structured input for output generation. In simple words, the structured input contains three parts: (i) the original query itself, (ii) the knowledge statement confirmed earlier, and (iii) the instruction to LLM on how to use the knowledge statement to answer the query. 

Selective contextual reasoning capitalizes on the emergent capabilities of LLMs to produce responses that are accurate, context-aware, and aligned with the latest knowledge. 
By decoupling the semantic filtering, knowledge confirmation, and contextual reasoning steps, it ensures efficiency, interpretability, and adaptability in sequential knowledge updating.
This method also enables the LLM to leverage its extensive knowledge base without the need for extensive fine-tuning or re-training, making it an efficient and effective alternative.

%=====================================
\section{Evaluation of Selective Contextual Reasoning}
\label{sec:assessment2}
%=====================================

Our experiments and detail analysis aim to address the following research questions:  \textbf{RQ1:} How does the performance of our proposed method SCR compare to existing model editing methods? \textbf{RQ2:} How robust is SCR to a varying number of knowledge updates? and \textbf{RQ3:} How does the retriever and top-$k$ retrieval impact the performance of knowledge updating?

%===============================
\subsection{Experimental Setup}
%===============================

Consistent with Section~\ref{sec:assessment}, we evaluate SCR against existing model editing methods on two datasets: WikiData\(_\text{counterfact}\) and ZsRE. 

\subsubsection{Backbone LLMs}
We conduct our experiments primarily on decoder-only LLMs, focusing specifically on Llama-2-7B-Chat \cite{touvron2023llama}, Llama-3.1-8B-Instruct \cite{llama3modelcard}, and the Mistral-7B-Instruct \cite{jiang2023mistral}, all of which are state-of-the-art autoregressive large language models.

Llama-2-7B-chat \cite{touvron2023llama} has 7 billion parameters, uses a decoder-only architecture, incorporating Grouped Query Attention (GQA) to optimize cache usage, reducing memory and computation. Llama-3.1-8B-Instruct \cite{llama3modelcard} has 8 billion parameters and is trained on over 15 trillion tokens, which is more than seven times the dataset used for Llama-2, Llama-3.1-8B outperforms Llama-2-7B across various benchmarks. 
Mistral-7B-Instruct \cite{jiang2023mistral} employs sliding window attention to extend context length and incorporates GQA to accelerate inference speed.

\subsubsection{Retriever}
For retriever, we employ \textbf{Contriever-msmarco}, a dense bi-encoder retriever fine-tuned on the MS-MARCO dataset, built upon the Contriever \cite{izacardunsupervised} model. In this architecture, the query and documents are encoded separately. 
The retriever model utilizes contrastive learning for unsupervised training, optimizing its representations through the InfoNCE loss function. Additionally, it incorporates data augmentation through an Inverse Cloze Task (ICT), thereby enhancing its training efficacy and robustness. 
In addition, we also use \textbf{Contriever} \cite{izacardunsupervised} and \textbf{ANCE} \cite{xiongapproximate} to study the impact of retriever choice on our evaluation. Contriever is the base model of Contriever-msmarco, with the same architecture and training objectives. ANCE retriever improves retrieval performance by leveraging an approximate nearest neighbor (ANN) index to efficiently sample high-quality hard negatives for contrastive learning. By replacing traditional random or in-batch negatives, ANCE enhances the model's ability to distinguish between similar yet irrelevant documents.
For all retrievers, relevance scores are calculated based on the dot product of their vector representations.

\subsubsection{Prompts}

There are two prompts in SCR: the \texttt{prompt} used for knowledge confirmation to determine the knowledge scope, and the \texttt{prompt} for in-context reasoning.

\begin{tcolorbox}[colback=white, colframe=black, title=Prompt for Knowledge Confirmation: , width=\textwidth]
\begin{minipage}{\textwidth}
Given a set of facts and a question, return the fact that best matches the core knowledge asked in the question. If the question cannot be answered with the facts, return "no relevant fact."\\

Facts: \{semantic-related knowledge\}\\
Question: \{question\}\\

Output:
\end{minipage}
\end{tcolorbox}

\begin{tcolorbox}[colback=white, colframe=black, title=Prompt for Contextual Reasoning:, width=\textwidth]
\begin{minipage}{\textwidth}
Answer the question based on the Updated Fact provided, without any explanation. At times, you need to think about the relationship between problems and facts before reasoning based on the information to answer.\\

Updated Fact: \{in-context knowledge\}\\
Question: \{question\}\\
Answer:
\end{minipage}
\end{tcolorbox}

%=================================
\subsection{Performance Comparisons}
\label{result_comparison}
%=================================

We evaluate SCR against ten existing model editing methods in lifelong knowledge updating scenario on two benchmark datasets and three backbone LLMs. Notably, we exclude results for MEMIT and PMET on Llama-3.1 and AlphaEdit on Mistral due to out-of-memory issues. The results are reported in Table~\ref{tab:result_comparasion}, with the best scores for each metric highlighted in bold.

\subsubsection{Performance on Llama-2}

Starting with WikiData\(_\text{counterfact}\), the best-performing baseline is GRACE, which excels in reliability and locality, achieving nearly 100. However, it shows limited performance in generalization and portability, scoring only 0.36 and 6.87, respectively. A more detailed analysis of the baseline performance can be found in Section~\ref{sec:assessment}. 
In contrast, SCR achieves the hightest scores in generalization and portability, with values of 53.28 and 25.67, respectively. Meanwhile, it maintains a reliability score of 52.32 and excels in locality, reaching 93.88. Overall, it outperforms other approaches in average effectiveness, attaining 56.29. By explicitly presenting the most relevant knowledge text, the LLM can more effectively capture the associations between different pieces of facts.

On the ZsRE dataset, our method continues to demonstrate strong performance across all metrics. It achieves remarkable results in generalization and portability, with scores of 65.18 and 41.13, respectively, which significantly outperform the second-best results of 24.44 and 9.74. Additionally, it maintains a relative high level of reliability of 79.78 and achieves outstanding locality with a score of 99.52. Our proposed SCR yields an impressive overall average score of 71.40, significantly surpassing all baseline model editing methods. 
While GRACE and WISE exhibit strong performance in specific metrics but struggle in others, our method displays a superior balance in success of knowledge updating and boundary delineation.

\subsubsection{Performance on Llama-3.1}
Turning to Llama-3.1, we observe similar outcomes, with our method generally outperforming baselines, particularly in generalization and portability. 

Specifically, on the WikiData\(_\text{counterfact}\) dataset, our method achieves a generalization score of 83.55, surpassing all baselines. Notably, portability remains strong at 35.29, significantly surpassing the next best-performing method by 154.99\%, indicating that updates made by our method transfer effectively across different contexts.
AlphaEdit demonstrates a balanced trade-off among all model editing methods, it achieves moderate scores across of 58.52, 42.43, 16.88, and 13.84 across the four evaluation metrics. However, its performance on Llama-2 is zero. This is because AlphaEdit's optimization process is highly sensitive to hyperparameters, which were specifically tuned for Llama-3 rather than Llama-2 in the original implementation.
On the ZsRE dataset, our method once again leads in generalization with a score of 66.56 and portability with 41.29, while maintaining relatively strong performance in reliability at 78.86 and locality at 98.74, second only to GRACE's 97.38 and 100, resulting in an overall average of 71.36.

\subsubsection{Performance on Mistral}
To further validate the effectiveness of SCR, we extend the knowledge updating LLM to Mistral-7B.
It is evident that after consecutive updates, methods like ROME and MEMIT almost entirely fail to retain the newly introduced knowledge. Moreover, the complete collapse of locality underscores the severe internal destabilization caused by localized parameter modifications. This disruption leads to a substantial loss of existing knowledge within the LLM. This suggests that such model editing methods are not only ineffective but may also introduce adverse effects. Among model editing baselines, WISE ranks second, with scores of 25.17 and 30.14 on the two datasets. GRACE emerges as the best-performing baseline, achieving scores of 50.13 and 51.29, thanks to its radius-based mechanism that intercepts hidden state inconsistencies. But they still fall short of broader reasoning capabilities.
Instead, our method outperforms all baselines, reaching leading average scores of 76.73 and 78.73. This result provides additional evidence for SCR's effectiveness in updating knowledge while preserving and transferring broader reasoning capabilities.

\subsubsection{Overall Analysis} 

The results clearly demonstrate that our method SCR consistently outperforms existing model editing methods on an overall level. 
It exhibits robust performance across different datasets and backbone LLMs, effectively achieving knowledge updating, knowledge isolation, and knowledge extension.
Moreover, SCR's performance improves as the reasoning and comprehension capabilities of LLMs advance. 
When evaluated on datasets with different levels of complexity, our method maintains stable and effective performance. Notably, WikiData\(_\text{counterfact}\) imposes stricter locality constraints compared to ZsRE, which is reflected in the forgetfulness metric in Section \ref{sec:assessment}. The dataset requires the LLM to recall original factual knowledge while excluding updated information. This increases the risk of performance fluctuations in model editing methods such as WISE that rely solely on token-level retrieval or surface-level pattern matching. Our method effectively mitigates these challenges through internal semantic understanding of LLMs.
Additionally, our method offers a more adaptable and balanced alternative to knowledge updating, excelling in generalization, portability, and overall performance. Unlike other methods that often compromise one dimension for another, our method provides a comprehensive solution that covers all crucial dimensions without sacrificing any. 
Our method effectively leverages the capabilities of different LLMs, adopting to their strengths for optimal performance. For example, on the challenging WikiData\(_\text{counterfact}\) dataset, the performance on Llama-3.1 clearly outperforms that of Llama-2.

The core innovation behind our method is the use of contextual reasoning, coupled with a dynamic and expandable memory framework that enables continuous knowledge updates. It empowers the LLM to dynamically incorporate recent and relevant knowledge into its inputs, rather than relying on static memorization. By adapting outputs based on the specific context provided by the input, our method allows the LLM to remain highly efficient, avoiding the need for expensive fine-tuning or re-training. 
In-context knowledge updating is highly efficient because it allows the LLM to utilize its extensive pre-trained knowledge while seamlessly incorporating new information. 
Ultimately, our method offers a computationally efficient and highly scalable method for knowledge updating.

\begin{table}
  \centering
  \caption{Performance Comparison on WikiData\(_\text{counterfact}\) and ZsRE Datasets. The backbone LLMs used for knowledge updating are Llama-2, Llama-3.1 and Mistral. Best results are highlighted in \textbf{bold}.}
  {\small
    \begin{tabular}{c|rrrrr|rrrrr}
    \toprule
    Dataset& \multicolumn{5}{c|}{\textbf{WikiData\(_\text{counterfact}\)}} & \multicolumn{5}{c}{\textbf{ZsRE}} \\
    Measure & \textbf{Rel.} & \textbf{Gen.} & \textbf{Loc.} & \textbf{Port.} & \textbf{Avg.} & \textbf{Rel.} & \textbf{Gen.} & \textbf{Loc.} & \textbf{Port.} & \textbf{Avg.} \\
    \midrule
    Method & \multicolumn{10}{l}{\textit{\textbf{Backbone model: Llama-2}}} \\
    \midrule
    Pre-edit & 0.12  & 0.36  & 100.00  & 6.82  & 26.83  & 1.85  & 2.23  & 100.00  & 4.58  & 27.17  \\
    ROME  & 0.00  & 0.00  & 0.00  & 0.00  & 0.00  & 2.28  & 1.69  & 0.48  & 0.76  & 1.30  \\
    MEMIT & 0.00  & 0.00  & 0.00  & 0.00  & 0.00  & 0.00  & 0.00  & 0.00  & 0.00  & 0.00  \\
    PMET  & 0.00  & 0.00  & 0.00  & 0.00  & 0.00  & 0.00  & 0.00  & 0.00  & 0.00  & 0.00  \\
    RECT  & 0.00  & 0.00  & 0.04  & 0.09  & 0.03  & 3.77  & 3.38  & 1.13  & 2.07  & 2.59  \\
    AlphaEdit & 0.00  & 0.00  & 0.00  & 0.00  & 0.00  & 0.00  & 0.00  & 0.00  & 0.00  & 0.00  \\
    MEND  & 0.00  & 0.00  & 0.00  & 0.00  & 0.00  & 0.00  & 0.00  & 0.00  & 0.00  & 0.00  \\
    FT-L  & 1.19  & 1.55  & 0.69  & 0.33  & 0.94  & 4.50  & 4.53  & 15.27  & 0.38  & 6.17  \\
    AdaLoRA & 0.00  & 0.00  & 0.48  & 0.50  & 0.25  & 0.08  & 0.08  & 0.00  & 0.00  & 0.04  \\
    GRACE & \textbf{91.90} & 0.36  & \textbf{98.29} & 6.87  & 49.35  & \textbf{92.77} & 2.77  & \textbf{100.00} & 4.59  & 50.03  \\
    WISE  & 31.35  & 20.74  & 72.72  & 12.01  & 34.20  & 30.05  & 24.44  & 91.80  & 9.74  & 39.01  \\
    \midrule
    \textbf{SCR}  & 52.32  & \textbf{53.28} & 93.88  & \textbf{25.67} & \textbf{56.29} & 79.78  & \textbf{65.18} & 99.52  & \textbf{41.13} & \textbf{71.40} \\
    \midrule
    Method & \multicolumn{10}{l}{\textit{\textbf{Backbone model: Llama-3.1}}} \\
    \midrule
    Pre-edit & 0.12  & 0.12  & 100.00  & 7.89  & 27.03  & 3.15  & 3.07  & 100.00  & 7.66  & 28.47  \\
    ROME  & 0.12  & 0.24  & 1.54  & 0.60  & 0.62  & 0.38  & 0.15  & 0.00  & 0.51  & 0.26  \\
    RECT  & 2.98  & 2.50  & 4.43  & 0.89  & 2.70  & 4.92  & 3.69  & 0.00  & 1.01  & 2.41  \\
    AlphaEdit & 58.52  & 42.43  & 16.88  & 13.84  & 32.92  & 77.25  & 65.18  & 76.42  & 13.65  & 58.12  \\
    MEND  & 0.00  & 0.00  & 0.00  & 0.00  & 0.00  & 0.00  & 0.00  & 0.00  & 0.00  & 0.00  \\
    FT-L  & 2.74  & 3.46  & 2.67  & 1.92  & 2.70  & 11.76  & 11.30  & 3.23  & 4.36  & 7.66  \\
    AdaLoRA & 0.36  & 0.36  & 10.65  & 2.36  & 3.43  & 0.08  & 0.08  & 0.00  & 0.00  & 0.04  \\
    GRACE & \textbf{95.35} & 0.12  & \textbf{98.44} & 8.02  & 50.48  & \textbf{97.38} & 3.69  & \textbf{100.00} & 7.66  & 52.18  \\
    WISE  & 1.91  & 2.38  & 88.72  & 6.42  & 24.86  & 13.60  & 13.22  & 91.53  & 6.17  & 31.13  \\
    \midrule
    \textbf{SCR}  & 92.37  & \textbf{83.55} & 97.96  & \textbf{35.29} & \textbf{77.29} & 78.86  & \textbf{66.56} & 98.74  & \textbf{41.29} & \textbf{71.36} \\
    \midrule
    Method & \multicolumn{10}{l}{\textit{\textbf{Backbone model: Mistral}}} \\
    \midrule
    Pre-edit & 0.12  & 0.12  & 100.00  & 6.90  & 26.79  & 2.69  & 2.38  & 100.00  & 5.65  & 27.68  \\
    ROME  & 0.00  & 0.12  & 1.22  & 0.08  & 0.36  & 0.92  & 0.77  & 0.00  & 0.51  & 0.55  \\
    MEMIT & 0.00  & 0.00  & 0.00  & 0.00  & 0.00  & 0.00  & 0.00  & 0.00  & 0.00  & 0.00  \\
    PEMT  & 0.00  & 0.00  & 0.00  & 0.00  & 0.00  & 0.00  & 0.00  & 0.00  & 0.00  & 0.00  \\
    RECT  & 0.95  & 0.72  & 0.23  & 0.84  & 0.69  & 1.84  & 2.15  & 0.25  & 1.27  & 1.38  \\
    MEND  & 0.00  & 0.00  & 0.34  & 0.04  & 0.10  & 0.00  & 0.00  & 0.00  & 0.00  & 0.00  \\
    FT-L  & 1.67  & 1.19  & 28.85  & 7.23  & 9.74  & 14.91  & 13.68  & \textbf{100.00}  & 5.04  & 33.41  \\
    AdaLoRA & 0.00  & 0.00  & 0.00  & 0.00  & 0.00  & 0.08  & 0.08  & 0.00  & 0.00  & 0.04  \\
    GRACE & \textbf{95.59} & 0.12  & \textbf{97.87} & 6.94  & 50.13  & \textbf{96.69} & 2.84  & \textbf{100.00} & 5.65  & 51.29  \\
    WISE  & 32.78  & 30.75  & 25.93  & 11.21  & 25.17  & 38.12  & 29.90  & 47.45  & 5.08  & 30.14  \\
    \midrule
    \textbf{SCR}  & 88.56  & \textbf{84.98} & 96.12  & \textbf{37.26} & \textbf{76.73} & 88.16  & \textbf{78.71} & 99.31  & \textbf{48.75} & \textbf{78.73} \\
    \bottomrule
    \end{tabular}
    }
  \label{tab:result_comparasion}
\end{table}

\subsection{Scaling to Different Number of Updates}
\label{different_nums_result}

Tables~\ref{tab:scale_to_100_10_llama2}, \ref{tab:scale_to_100_10_llama3.1}, and \ref{tab:scale_to_100_10_mistral} present the results of Llama-2, Llama-3.1, and Mistral respectively,  after 10 to 100 rounds in a lifelong knowledge updating scenario. This comparison highlights the impact of scaling on the effectiveness of different model editing techniques and our method.

\subsubsection{Analysis of Existing Model Editing Methods}

Model editing methods exhibit distinct behaviors when scaling the number of updates, which can be broadly categorized into three groups: those that are sensitive to the number of updates and experience significant degrade, those that remain resilient but have specific limitations, and those that remain ineffective regardless of the scale of updates.

As the number of updates increases from 10 to 100, we observe a general trend of performance degradation across most model editing methods.
This decline is particularly pronounced in methods such as ROME, MEMIT, PEFT, and RECT, which perform well with only 10 updates but exhibit significant drops when scaled to 100. 
For example, ROME achieves an impressive average of 44.79 for WikiData\(_\text{counterfact}\) and 46.39 for ZsRE with 10 updates on Llama-2. However, as the number reaches 100, its effectiveness diminishes to zero. 
Similarly, RECT achieves a average of 49.36 for WikiData\(_\text{counterfact}\) and 59.17 for ZsRE with 10 updates on Llama-3.1, but its performance noticeably deteriorates to 7.89 and 6.75 when the number is increased to 100.
Moreover, even with a minimal number of updates, the locality of these methods can drop below 10, leading to a complete model collapse. For instance, after just 10 updates on the WikiData\(_\text{counterfact}\) dataset, ROME reduces the locality score of Llama-3.1 to 9.58.
This suggests that while these methods can handle a limited number of updates with high precision, they struggle to retain effectiveness under the cumulative effects of numerous modifications. This degradation can be attributed to cumulative parameter shifts, which lead the LLM to gradually forget previously acquired information, including recent updates and original knowledge.

In contrast, certain methods exhibit greater resilience to an increasing number of updates. 
Among them, GRACE maintains high effectiveness even with 100 updates, achieving an average of 50.37 on WikiData\(_\text{counterfact}\) and 50.92 on ZsRE on Llama-2. 
And AlphaEdit also exhibits robust performance on Llama-3.1, starting with 41.00 on WikiData\(_\text{counterfact}\) and 59.17 on ZsRE at 10 updates and sustaining competitive scores of 48.48 and 62.92, respectively, at 100 updates.
Additionally, WISE demonstrates strong stability under large-scale edits, achieving 44.07 and 59.17 at 10 updates and maintaining 46.31 and 55.66, respectively, at 100 updates on Mistral.
While these methods demonstrate stability, they each have certain limitations. GRACE tends to rely heavily on strict pattern matching, meaning it performs well only when encountering identical queries but struggles with variations. Similarly, AlphaEdit is not well-suited for reasoning over updated knowledge. WISE, on the other hand, has difficulty accurately delineating the boundaries of newly integrated knowledge.

Meanwhile, some methods remain ineffective regardless of the update scale, including MEND, FT-L, and AdaLoRA. MEND, for instance, consistently delivers poor results, failing to achieve meaningful improvements even with a small number of updates. This lack of adaptability makes these methods ill-suitable for scenarios that require frequent updates or modifications, as they struggle to integrate new information effectively.
MEND relies on a hypernetwork to generate updates for model weights, but this mechanism is highly sensitive to changes in the parameter distribution of LLMs. In the case of frequent updates, FT-L and AdaLoRA may disrupt the LLM’s original knowledge representations, making it difficult for new and old knowledge to coexist.

\subsubsection{Observation on SCR}
Our method demonstrates superiority across multiple dimensions while maintaining strong robustness to the number of updates.
Specifically, when performing knowledge updates on Llama-3.1, the recall performance across different knowledge formats (generalization) remains around 90, while knowledge utilization (portability) reaches approximately 40.
Moreover, SCR demonstrates exceptional resilience to update frequency. On Mistral, regardless of whether knowledge is updated 10 or 100 rounds, reliability remains consistently above 90.
Unlike methods that struggle with large-scale knowledge updates, SCR dynamically adjusts its search space, ensuring that the update process remains precise and controlled.
Regardless of the size of the knowledge base, SCR's two-step selection strategy consistently narrows the knowledge comparison and verification space to a minimal range.
In the first step, only the most relevant knowledge candidates are retained, significantly reducing unnecessary interference. In the second step, the LLM leverages its own reasoning capabilities to further prune the selection, discarding less useful information and preserving only the most essential knowledge for the task at hand.
Additionally, instead of treating each knowledge update as an isolated fact, SCR analyzes the context surrounding the new information within the LLM, making full use of the relationship between new knowledge and other original knowledge. In addition, contextual reasoning enables SCR to resolve inconsistencies between old and new knowledge, prioritizing the most recent one.

\begin{table}
  \centering
  \caption{The number of sequential knowledge updates changes from 10 to 100, with Llama-2.}
  \resizebox{\textwidth}{!}{
    \begin{tabular}{cl|rrrrr|rrrrr}
    \toprule
    \multicolumn{2}{c|}{Dataset $\Rightarrow$}& \multicolumn{5}{|c|}{\textbf{WikiData\(_\text{counterfact}\)}} & \multicolumn{5}{c}{\textbf{ZsRE}} \\
    \textbf{\#Editing} &\textbf{Method}     & \textbf{Rel.} & \textbf{Gen.} & \textbf{Loc.} & \textbf{Port.} & \textbf{Avg.} & \textbf{Rel.} & \textbf{Gen.} & \textbf{Loc.} & \textbf{Port.} & \textbf{Avg.} \\
    \midrule
    
    \multirow{12}[4]{*}{10} & Pre-edit & 0.00  & 0.00  & 100.00  & 2.65  & 25.66  & 0.00  & 0.00  & 100.00  & 0.00  & 25.00  \\
          & ROME  & \textbf{100.00 } & 30.00  & 26.36  & 22.81  & 44.79  & \textbf{100.00 } & 50.00  & 22.22  & 13.33  & 46.39  \\
          & MEMIT & 90.00  & 40.00  & 7.10  & \textbf{37.34 } & 43.61  & 90.00  & 50.00  & 44.44  & 6.67  & 47.78  \\
          & PMET  & 0.00  & 0.00  & 54.74  & 4.83  & 14.89  & 0.00  & 0.00  & 88.89  & 6.67  & 23.89  \\
          & RECT  & \textbf{100.00 } & 40.00  & 31.69  & 25.65  & 49.33  & 90.00  & 60.00  & 11.11  & 6.67  & 41.94  \\
          & AlphaEdit & 0.00  & 0.00  & 0.00  & 0.00  & 0.00  & 0.00  & 0.00  & 0.00  & 0.00  & 0.00  \\
          & MEND  & 30.00  & 10.00  & 16.65  & 11.90  & 17.14  & 0.00  & 0.00  & 0.00  & 0.00  & 0.00  \\
          & FT-L  & 0.00  & 10.00  & 0.00  & 0.00  & 2.50  & 20.00  & 30.00  & 33.33  & 0.00  & 20.83  \\
          & AdaLoRA & 10.00  & 0.00  & 0.00  & 0.00  & 2.50  & 10.00  & 10.00  & 0.00  & 0.00  & 5.00  \\
          & GRACE & 80.00  & 0.00  & \textbf{100.00 } & 2.65  & 45.66  & \textbf{100.00 } & 0.00  & \textbf{100.00 } & 0.00  & 50.00  \\
          & WISE  & 80.00  & 60.00  & 35.38  & 32.34  & 51.93  & \textbf{100.00 } & \textbf{100.00 } & 22.22  & 16.67  & 59.72  \\
\cmidrule{2-12}          & \textbf{SCR}  & 60.00  & \textbf{70.00 } & 93.37  & 37.23  & \textbf{65.15 } & 70.00  & 60.00  & 88.89  & \textbf{50.00 } & \textbf{67.22 } \\
    \midrule
    \multirow{12}[4]{*}{100} & Pre-edit & 0.00  & 1.00  & 100.00  & 5.73  & 26.68  & 2.00  & 3.00  & 100.00  & 3.66  & 27.17  \\
          & ROME  & 0.00  & 0.00  & 0.00  & 0.00  & 0.00  & 10.00  & 10.00  & 0.00  & 1.33  & 5.33  \\
          & MEMIT & 0.00  & 0.00  & 0.00  & 0.00  & 0.00  & 52.00  & 42.00  & 21.05  & 8.55  & 30.90  \\
          & PMET  & 0.00  & 0.00  & 1.83  & 0.10  & 0.48  & 3.00  & 2.00  & 78.95  & 1.26  & 21.30  \\
          & RECT  & 0.00  & 0.00  & 0.20  & 0.14  & 0.09  & 17.00  & 17.00  & 7.89  & 0.63  & \multicolumn{1}{c}{10.63 } \\
          & AlphaEdit & 0.00  & 0.00  & 0.00  & 0.00  & 0.00  & 0.00  & 0.00  & 0.00  & 0.00  & 0.00  \\
          & MEND  & 0.00  & 0.00  & 0.00  & 0.00  & 0.00  & 0.00  & 0.00  & 0.00  & 0.00  & 0.00  \\
          & FT-L  & 1.00  & 2.00  & 0.00  & 0.00  & 0.75  & 17.00  & 18.00  & 5.26  & 0.63  & 10.22  \\
          & AdaLoRA & 0.00  & 0.00  & 5.65  & 0.90  & 1.64  & 0.00  & 0.00  & 0.00  & 0.00  & 0.00  \\
          & GRACE & \textbf{95.00 } & 1.00  & \textbf{99.75 } & 5.73  & 50.37  & \textbf{97.00 } & 3.00  & \textbf{100.00 } & 3.66  & 50.92  \\
          & WISE  & 87.00  & 52.00  & 57.48  & 27.48  & 55.99  & 51.00  & 34.00  & 76.32  & 7.48  & 42.20  \\
\cmidrule{2-12}          & \textbf{SCR}  & 57.00  & \textbf{62.00 } & 93.85  & \textbf{30.37 } & \textbf{60.80 } & 81.00  & \textbf{73.00 } & 97.37  & \textbf{39.26 } & \textbf{72.66 } \\

    \bottomrule
    
    \end{tabular}
}
\label{tab:scale_to_100_10_llama2}
\end{table}

\begin{table}
  \centering
  \caption{The number of sequential knowledge changes from 10 to 100, with Llama-3.1.}
  \resizebox{\textwidth}{!}{
    \begin{tabular}{cl|rrrrr|rrrrr}
    \toprule
    \multicolumn{2}{c|}{Dataset $\Rightarrow$}& \multicolumn{5}{|c|}{\textbf{WikiData\(_\text{counterfact}\)}} & \multicolumn{5}{c}{\textbf{ZsRE}} \\
    \textbf{\#Editing} &\textbf{Method}    & \textbf{Rel.} & \textbf{Gen.} & \textbf{Loc.} & \textbf{Port.} & \textbf{Avg.} & \textbf{Rel.} & \textbf{Gen.} & \textbf{Loc.} & \textbf{Port.} & \textbf{Avg.} \\
    \midrule
      \multirow{10}[4]{*}{10} & Pre-edit & 0.00  & 0.00  & 100.00  & 3.76  & 25.94  & 0.00  & 0.00  & 100.00  & 6.67  & 26.67  \\
          & ROME  & 50.00  & 30.00  & 9.58  & 19.59  & 27.29  & 80.00  & 60.00  & 50.00  & 30.00  & 55.00  \\
          & RECT  & 90.00  & 70.00  & 12.04  & 25.40  & 49.36  & 90.00  & 50.00  & 66.67  & 30.00  & 59.17  \\
          & AlphaEdit & 90.00  & 10.00  & 43.03  & 20.99  & 41.00  & \textbf{100.00 } & 30.00  & \textbf{100.00 } & 6.67  & 59.17  \\
          & MEND  & 0.00  & 0.00  & 0.00  & 0.00  & 0.00  & 0.00  & 0.00  & 0.00  & 0.00  & 0.00  \\
          & FT-L  & 0.00  & 0.00  & 0.00  & 0.00  & 0.00  & 20.00  & 20.00  & 0.00  & 16.67  & 14.17  \\
          & AdaLoRA & 10.00  & 0.00  & 0.00  & 0.00  & 2.50  & 10.00  & 10.00  & 0.00  & 0.00  & 5.00  \\
          & GRACE & \textbf{100.00 } & 0.00  & \textbf{100.00 } & 3.76  & 50.94  & \textbf{100.00 } & 0.00  & \textbf{100.00 } & 6.67  & 51.67  \\
          & WISE  & 60.00  & 50.00  & 29.83  & 17.02  & 39.21  & 50.00  & 50.00  & 33.33  & 17.78  & 37.78  \\
\cmidrule{2-12}          & \textbf{SCR}  & \textbf{100.00 } & \textbf{90.00 } & 94.69  & \textbf{31.29 } & \textbf{78.99 } & 90.00  & \textbf{100.00 } & 83.33  & \textbf{53.33 } & \textbf{81.67 } \\
    \midrule
    \multirow{10}[4]{*}{100} & Pre-edit & 0.00  & 0.00  & 100.00  & 7.92  & 26.98  & 5.00  & 3.00  & 100.00  & 6.88  & 28.72  \\
          & ROME  & 5.00  & 5.00  & 13.38  & 1.17  & 6.14  & 7.00  & 6.00  & 0.00  & 2.67  & 3.92  \\
          & RECT  & 3.00  & 7.00  & 19.33  & 2.24  & 7.89  & 11.00  & 12.00  & 0.00  & 4.00  & 6.75  \\
          & AlphaEdit & 86.00  & 44.00  & 42.17  & 21.76  & 48.48  & 89.00  & 71.00  & 80.95  & 10.73  & 62.92  \\
          & MEND  & 0.00  & 0.00  & 0.00  & 1.28  & 0.32  & 0.00  & 0.00  & 0.00  & 0.00  & 0.00  \\
          & FT-L  & 0.00  & 0.00  & 0.00  & 1.28  & 0.32  & 10.00  & 11.00  & 0.00  & 5.33  & 6.58  \\
          & AdaLoRA & 1.00  & 0.00  & 5.57  & 1.76  & 2.08  & 1.00  & 0.00  & 0.00  & 0.00  & 0.25  \\
          & GRACE & \textbf{99.00 } & 0.00  & \textbf{99.84 } & 7.94  & 51.69  & \textbf{100.00 } & 3.00  & \textbf{100.00 } & 6.88  & 52.47  \\
          & WISE  & 6.00  & 5.00  & 75.33  & 10.24  & 24.14  & 27.00  & 19.00  & 78.57  & 9.28  & 33.46  \\
\cmidrule{2-12}          & \textbf{SCR}  & 97.00  & \textbf{91.00 } & 99.16  & \textbf{37.88 } & \textbf{81.26 } & 86.00  & \textbf{81.00 } & 97.62  & \textbf{43.11 } & \textbf{76.93 } \\
    
    \bottomrule
    \end{tabular}
}
\label{tab:scale_to_100_10_llama3.1}
\end{table}

\begin{table}
  \centering
  \caption{The number of sequential knowledge updates changes from 10 to 100, with Mistral.}
  \resizebox{\textwidth}{!}{
    \begin{tabular}{cl|ccccc|ccccc}
    \toprule
    \multicolumn{2}{c|}{Dataset $\Rightarrow$}& \multicolumn{5}{|c|}{\textbf{WikiData\(_\text{counterfact}\)}} & \multicolumn{5}{c}{\textbf{ZsRE}} \\
     \textbf{\#Editing} &\textbf{Method}     & \textbf{Rel.} & \textbf{Gen.} & \textbf{Loc.} & \textbf{Port.} & \textbf{Avg.} & \textbf{Rel.} & \textbf{Gen.} & \textbf{Loc.} & \textbf{Port.} & \textbf{Avg.} \\
    \midrule
    \multirow{9}[4]{*}{10} & Pre-edit & 0.00  & 0.00  & 35.33  & 1.11  & 9.11  & 0.00  & 0.00  & 25.00  & 0.00  & 0.00  \\
          & ROME  & 80.00  & 50.00  & 15.61  & 33.03  & 44.66  & 80.00  & 70.00  & \textbf{100.00 } & 6.67  & 64.17  \\
          & RECT  & 90.00  & 50.00  & 26.22  & 25.40  & 47.91  & \textbf{100.00 } & 70.00  & \textbf{100.00 } & 6.67  & 69.17  \\
          & MEND  & 0.00  & 0.00  & 0.00  & 0.79  & 0.20  & 0.00  & 0.00  & 0.00  & 0.00  & 0.00  \\
          & FT-L  & 0.00  & 0.00  & 67.19  & 0.00  & 16.80  & 10.00  & 0.00  & \textbf{100.00 } & 0.00  & 27.50  \\
          & AdaLoRA & 0.00  & 0.00  & 0.00  & 0.00  & 0.00  & 10.00  & 10.00  & 0.00  & 0.00  & 5.00  \\
          & GRACE & \textbf{100.00 } & 0.00  & \textbf{100.00 } & 1.11  & 50.28  & \textbf{100.00 } & 0.00  & \textbf{100.00 } & 0.00  & 50.00  \\
          & WISE  & 50.00  & 70.00  & 35.14  & 21.13  & 44.07  & 90.00  & 90.00  & 40.00  & 16.67  & 59.17  \\
\cmidrule{2-12}          & \textbf{SCR}  & \textbf{100.00 } & \textbf{100.00 } & 98.83  & \textbf{42.03 } & \textbf{85.21 } & \textbf{100.00 } & \textbf{100.00 } & 80.00  & \textbf{66.67 } & \textbf{86.67 } \\
    \midrule
    \multirow{9}[4]{*}{100} & Pre-edit & 0.00  & 0.00  & 38.54  & 5.97  & 11.13  & 2.00  & 3.00  & 16.50  & 4.66  & 6.54  \\
          & ROME  & 3.00  & 2.00  & 30.18  & 0.99  & 7.48  & 5.00  & 4.00  & 0.00  & 0.00  & 2.25  \\
          & RECT  & 6.00  & 4.00  & 18.10  & 1.81  & 7.48  & 10.00  & 9.00  & 0.00  & 1.33  & 5.08  \\
          & MEND  & 0.00  & 0.00  & 0.26  & 0.18  & 0.11  & 0.00  & 0.00  & 0.00  & 0.00  & 0.00  \\
          & FT-L  & 3.00  & 2.00  & 65.07  & 8.51  & 19.64  & 17.00  & 17.00  & 93.94  & 6.25  & 33.55  \\
          & AdaLoRA & 1.00  & 0.00  & 7.20  & 1.76  & 2.49  & 1.00  & 0.00  & 0.00  & 0.00  & 0.25  \\
          & GRACE & \textbf{100.00 } & 0.00  & \textbf{99.78 } & 5.97  & 51.44  & \textbf{98.00 } & 3.00  & \textbf{100.00 } & 4.66  & 51.42  \\
          & WISE  & 75.00  & 70.00  & 14.22  & 26.03  & 46.31  & 71.00  & 61.00  & 78.79  & 11.85  & 55.66  \\
\cmidrule{2-12}          & \textbf{SCR}  & 92.00  & \textbf{97.00 } & 97.81  & \textbf{43.62 } & \textbf{82.61 } & 93.00  & \textbf{88.00 } & 93.94  & \textbf{45.74 } & \textbf{80.17 } \\
    
    \bottomrule
    \end{tabular}
  }
  \label{tab:scale_to_100_10_mistral}
\end{table}

%=====================================
\subsection{Hyper-parameter Experiments} 
%=====================================
In this section, we study the impact of the number of retrieved knowledge statements, and the retriever.

\subsubsection{Effect of top-$k$ Retrieval}
\begin{figure}
	\centering
	\subfloat[Llama-2, WikiData\(_\text{counterfact}\)]{
	    \includegraphics[width=0.42\textwidth]{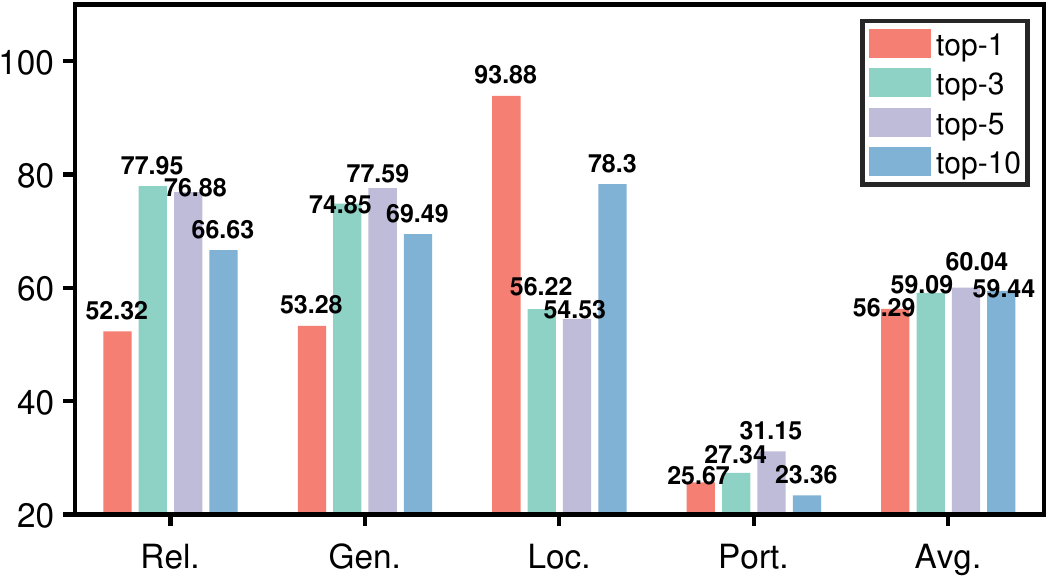}
	}
    \subfloat[Llama-2, ZsRE]{
	    \includegraphics[width=0.42\textwidth]{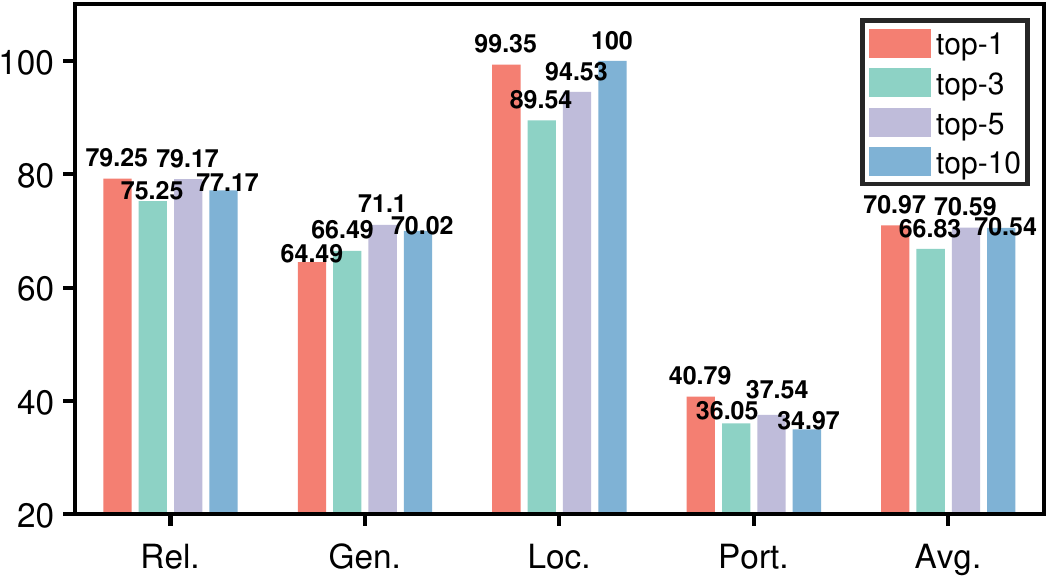}
	}\\
	\subfloat[Llama-3.1, WikiData\(_\text{counterfact}\)]{
	    \includegraphics[width=0.42\textwidth]{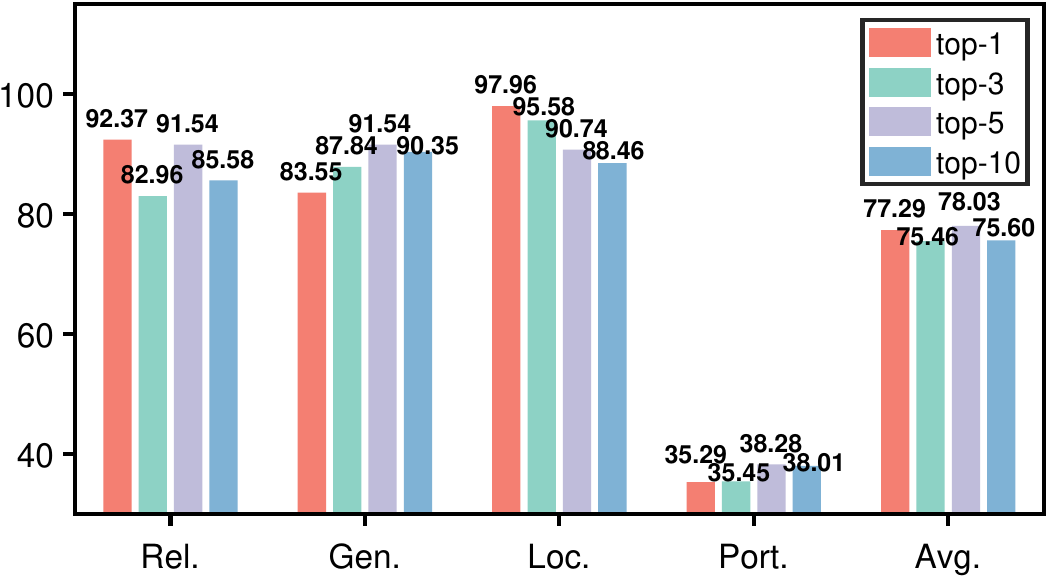}
	}
    \subfloat[Llama-3.1, ZsRE]{
	    \includegraphics[width=0.42\textwidth]{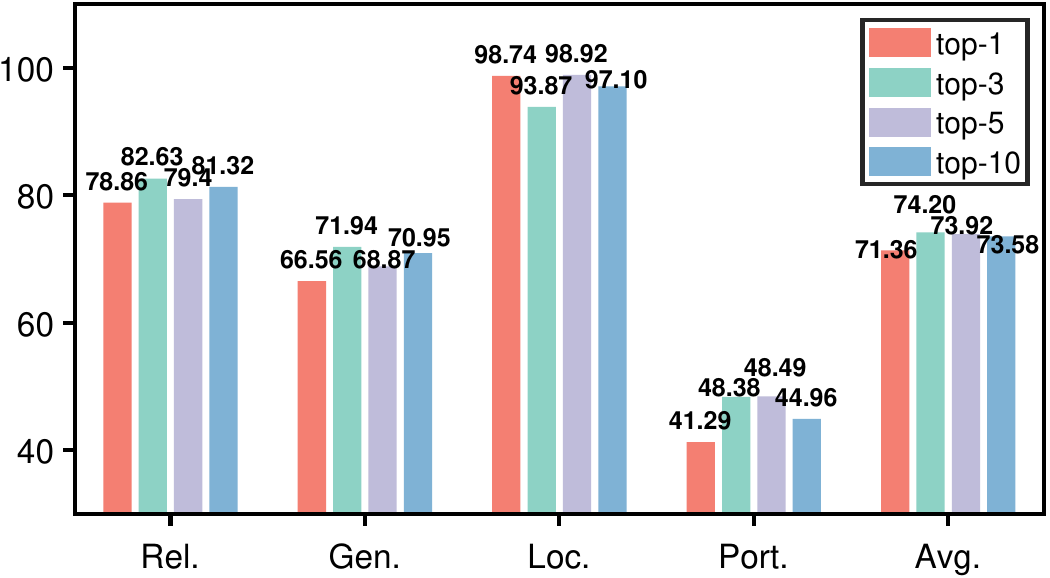}
	}\\
\subfloat[Mistral, WikiData\(_\text{counterfact}\)]{
	    \includegraphics[width=0.42\textwidth]{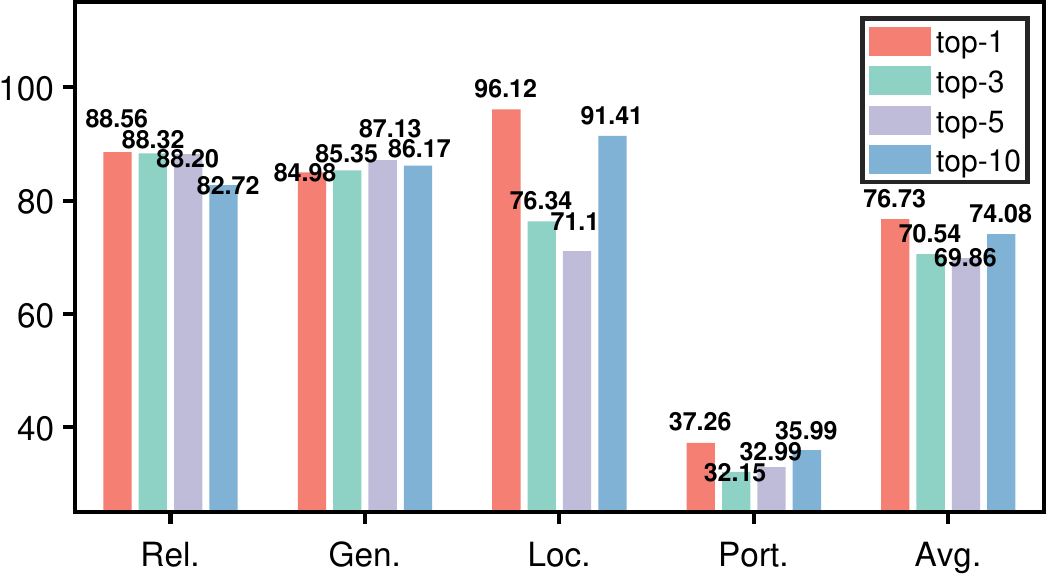}
	}
    \subfloat[Mistral, ZsRE]{
	    \includegraphics[width=0.42\textwidth]{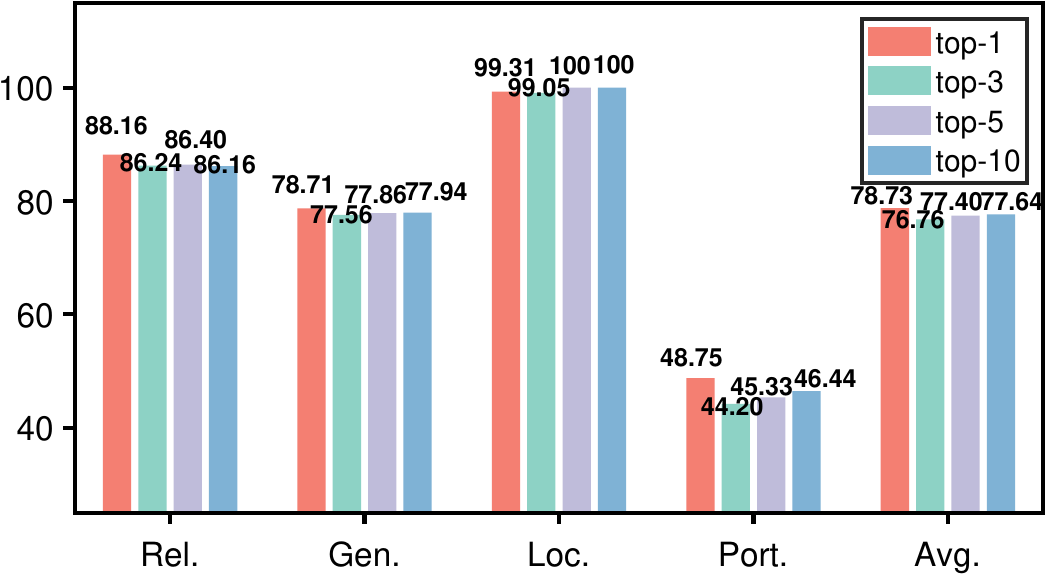}
	}
	\caption{Effect of top-$k$ selection in the semantic filtering step when updating knowledge in WikiData\(_\text{counterfact}\) and ZsRE with three backbone models: Llama-2, Llama-3.1, and Mistral.}
	\label{fig:3topk}
\end{figure}

One of the main advantages of our method is its simplicity and strong effectiveness. As shown in Section \ref{result_comparison} and Section \ref{different_nums_result}, our method achieves notable improvements over traditional model editing methods across different backbone LLMs and varying numbers of edits. On the other hand, different numbers of facts may have varying impacts on performance of knowledge confirmation. We perform additional experiments to analyze how the number of retrieved entries affects the evaluation metrics. Specifically, we test with top-1, top-3, top-5, and top-10 retrieval facts, and the results are presented in Figs.~\ref{fig:3topk}.

The effects across different datasets and backbone LLMs varies due to variations in data distribution and the capabilities of LLMs.
Updating Llama-2, WikiData\(_\text{counterfact}\) may benefit more from additional retrieved facts, while ZsRE may be more sensitive to the relevance and specificity of the top-1 fact. For WikiData\(_\text{counterfact}\), top-5 retrieval generally yields the best overall performance, but at the cost of significantly lower locality. In contrast, for ZsRE, top-1 retrieval consistently performs best, particularly in terms of reliability and portability. These results highlight a fundamental trade-off in knowledge updating: increasing the number of retrieved facts enhances generality but compromises portability. Updating Llama-3.1, on WikiData\(_\text{counterfact}\), top-5 retrieval achieves the best performance overall. For ZsRE, top-3 retrieval achieves the best results, presenting an inverse effect compared to the performance observed on Llama-2. This discrepancy suggests that model-specific retrieval dynamics play a crucial role in knowledge integration, potentially influenced by differences in pretraining corpus composition, parameterization, or architectural enhancements.
Updating Mistral, on WikiData\(_\text{counterfact}\), the performance is highly sensitive to the choice of $k$; however, for ZsRE, the results remain relatively consistent across different values of $k$.

Overall, retrieving more facts provides more comprehensive information but also introduces additional noise, making top-3 and top-5 retrievals balanced and reliable choices. 

\subsubsection{Effect of Retriever}
We then explore the impact of the retriever on knowledge updating performance. The primary role of the retriever is to select the most semantically relevant factual knowledge from the external memory in response to a given query.  To examine this effect, we evaluate three different retrievers, including Contriever-msmarco \cite{izacardunsupervised}, Contriever \cite{izacardunsupervised}, and ANCE \cite{xiongapproximate}, in knowledge updating  tasks for Llama-2, Llama-3.1, and Mistral, using the ZsRE dateset and top-1 retrieval strategy.

As shown in Table~\ref{tab:3retriever}, Contriever-msmarco, as a state-of-the-art dense retriever, achieves solid performance across all backbones, achieving scores of 70.97, 71.36, and 78.73, respectively. It consistently provides highly relevant context, thereby enhancing the accuracy and effectiveness.
Both Contriever and ANCE also deliver relatively strong results. However, the former exhibits a noticeable decline in reliability, with an average score 2.12 points lower than Contriever-msmarco. The latter, on the other hand, faces challenges with portability, scoring 2.28 points lower than Contriever-msmarco averagely.

In summary, the retriever plays a crucial role in knowledge updating by determining the relevance and quality of the retrieved context. A high-quality retriever like Contriever-msmarco can not only improve the relevance of retrieved facts, alleviating the burden of knowledge confirmation, but also further improve the overall success rate of knowledge integration. Furthermore, even when using the same retriever and retrieval strategy, the effectiveness of knowledge updates varies across different LLMs, once again underscoring the importance of the LLM’s inherent understanding and reasoning capabilities. In addition, future work can also explore more effective retrieval methods, such as confidence-based retrieval or similarity comparison based on LLMs' representations.

\begin{table}
  \centering
  \caption{Performance of retrievers for knowledge updating in three backbone models, on ZsRE dataset.}
  {\small 
    \begin{tabular}{c|c|ccccc}
    \toprule
\textbf{Backbone LLM} &   \textbf{Retriever} & \textbf{Rel.} & \textbf{Gen.} & \textbf{Loc.} & \textbf{Port.} & \textbf{Avg.} \\
    \midrule
 &    Contriever-msmarco & \textbf{79.25} & \textbf{64.49 }  &\textbf{99.35} & \textbf{40.79 } & \textbf{70.97} \\

 Llama-2  &   Contriever & 74.10  & 63.26  & 96.14  & 40.16  & 68.42  \\
   & ANCE  & 77.25  & 63.72  & 96.14  & 36.78  & 68.47  \\
    \midrule
    
  &  Contriever-msmarco & \textbf{78.86 } & \textbf{66.56 } & 98.74  & \textbf{41.29 } & \textbf{71.36 } \\
   Llama-3.1&  Contriever & 73.48  & 65.64  & \textbf{99.25 } & 40.38  & 69.69  \\
    &ANCE  & 76.33  & 65.87  & \textbf{99.25 } & 36.99  & 69.61  \\
        \midrule
     & Contriever-msmarco & \textbf{88.16 } & \textbf{78.71 } & \textbf{99.31 } & \textbf{48.75 } & \textbf{78.73 } \\
   Mistral& Contriever & 82.46  & 77.56  & 99.05  & 47.24  & 76.58  \\
    &ANCE  & 85.55  & 77.79  & 98.55  & 42.71  & 76.15 
\\
    \bottomrule
    \end{tabular}
    }
  \label{tab:3retriever}
\end{table}

\subsection{Limitations of Evaluation}
Due to the constraints of our available computing resources, our experiments have primarily centered on LLMs with fewer than 10 billion parameters. While this limits the scale of LLMs we could investigate, the results still demonstrate the effectiveness and robustness of our proposed method.
We believe, however, our method could yield even better performance when applied to larger LLMs. With their enhanced capacity for complex reasoning, deeper contextual understanding, and more nuanced knowledge representation, larger LLMs are likely to benefit even more from the selective contextual reasoning framework we propose. We are confident that our method will scale well, continuing to deliver significant improvements as the LLM size and capabilities grow.

%=================================
\section{Conclusion and Future Work}
%=================================
In this work, we have evaluated ten widely used model editing methods within the context of autoregressive generation for lifelong knowledge updating scenarios. Through a series of experiments, we have observed that although recent model editing techniques prove effective for a specific knowledge update, they exhibit a decline in performance as the number of updates increases.
Moreover, no recent model editing method achieves a comprehensive advantage across all four dimensions: reliability, generalization, locality, and portability.

In response, we propose a novel method called Selective Contextual Reasoning (SCR) that offers a more scalable and robust alternative. Instead of directly altering parameters of LLMs, our method dynamically retrieves and evaluates relevant knowledge from an external memory, ensuring both knowledge preservation and sustained performance, even with extensive updates. This selected knowledge is then contextualized and integrated into the inference process.

Our experimental results, on two datasets across three backbone LLMs, demonstrate that our proposed SCR consistently outperforms existing model editing methods across multiple datasets and backbone LLMs. Notably, the performance of our method improves significantly with advancements in the reasoning and comprehension abilities of LLMs, highlighting its scalability and robustness across various architectures and data complexities.
Additionally, we analyze the impact of key hyperparameters, such as the number of retrieved facts, and examine the role of different retrievers in influencing the quality of response. Our findings highlight that high-quality retrievers, like Contriever-msmarco, significantly enhance the relevance and accuracy of the selected knowledge context, thereby improving the performance of knowledge updating across various backbone LLMs.

This work highlights the limitations of recent model editing methods and emphasizes the potential of contextual reasoning for scalable and efficient lifelong learning of LLMs. Future research could focus on optimizing knowledge selection strategies and integrating more complex reasoning mechanisms to further enhance the capabilities of reasoning-augmented knowledge updating.